\RequirePackage[svgnames]{xcolor}

\documentclass[11pt,letterpaper]{mystyle}
\usepackage[all]{hypcap}
\usepackage[svgnames]{xcolor}
\usepackage[comma,authoryear,compress]{natbib}
\usepackage{hyperref}[citecolor=lightblue]
\usepackage{amsmath}
\usepackage{etoolbox}

\hypersetup{
    colorlinks = true,
    citecolor = {YaleBlue},
}

\usepackage{algorithm}
\usepackage{algorithmicx}
\usepackage{algpseudocode}
\usepackage{microtype}
\usepackage{graphicx}
\expandafter\def\csname ver@subfig.sty\endcsname{}
\usepackage{booktabs} %
\usepackage{float}
\usepackage{bigstrut}

\usepackage{comment}
\usepackage{amsmath}
\usepackage{amssymb}
\usepackage{mathtools}
\usepackage{amsthm}
\usepackage{mathrsfs}
\usepackage{nicefrac}
\usepackage{dsfont}
\usepackage{enumitem}
\usepackage{subcaption}
\usepackage{graphicx,subfig}
\usepackage{cleveref}
\usepackage{bxcoloremoji}
\usepackage{float}
\usepackage{algpseudocode}
\usepackage{xspace}

\usepackage[utf8]{inputenc} %
\usepackage[T1]{fontenc}    %
\usepackage{url}            %
\usepackage{booktabs}       %
\usepackage{amsfonts}       %
\usepackage{nicefrac}       %
\usepackage{microtype}      %
\usepackage{graphicx}
\usepackage{amssymb}
\usepackage{fdsymbol}
\usepackage{wrapfig}
\usepackage{lipsum}
\usepackage{enumitem}
\usepackage{stackengine}
\usepackage[font=small,labelfont=bf]{caption}
\usepackage{color}
\usepackage{adjustbox}

\usepackage{rotating}
\usepackage{makecell}

\usepackage{multirow}

\newcommand{\x}{\mathbf{x}}

\definecolor{blanchedalmond}{rgb}{1.0, 0.92, 0.8}
\definecolor{carmine}{rgb}{0.59, 0.0, 0.09}
\definecolor{lightblue}{rgb}{0.22,0.45,0.70}%

\renewcommand{\mathbf}{\boldsymbol}

\makeatletter
\def\Ddots{\mathinner{\mkern1mu\raise\p@
\vbox{\kern7\p@\hbox{.}}\mkern2mu
\raise4\p@\hbox{.}\mkern2mu\raise7\p@\hbox{.}\mkern1mu}}
\makeatother

\definecolor{amaranth}{rgb}{0.9, 0.17, 0.31}
\definecolor{antiquebrass}{rgb}{0.8, 0.58, 0.46}
\definecolor{antiquefuchsia}{rgb}{0.57, 0.36, 0.51}
\definecolor{chromeyellow}{rgb}{0.31, 0.47, 0.26}

\newtcolorbox{AIbox}[2][]{aibox,title=#2,#1}
\definecolor{lightblue}{rgb}{0.22,0.45,0.70}%
\definecolor{Gray}{gray}{0.95}
\definecolor{Cornsilk}{rgb}{1.0, 0.97, 0.86}

\usepackage{url}
\usepackage{pifont}
\usepackage{soul}

\usepackage{amsmath}

\usepackage[all]{hypcap}

\usepackage{algpseudocode}
\usepackage{authblk}   
\usepackage{amssymb}      
\usepackage{bbm}          
\usepackage{optidef}  

\usepackage[utf8]{inputenc}
\usepackage[T1]{fontenc}

\usepackage{mathpazo}
\usepackage{tcolorbox}
\tcbuselibrary{skins,breakable,listings} 

\usepackage{lipsum}

\newtcolorbox{simpleElegantQuote}{
    colback=AliceBlue!50!White,   
    colframe=RoyalBlue!75!Black,  
    boxrule=0.5pt,                
    arc=2mm,                     
    boxsep=4pt,                   
    left=10pt, right=10pt,        
    top=8pt, bottom=8pt,         
    fontupper=\itshape,          
}

\usepackage{etoolbox}
\usepackage{xspace}
\usepackage{amsmath}
\usepackage{enumitem}
\usepackage{multirow}
\usepackage{lineno}

\usepackage[utf8]{inputenc} 
\usepackage{url}            
\usepackage{booktabs}       
\usepackage{amsfonts}       
\usepackage{nicefrac}       
\usepackage{microtype}      

\usepackage{graphicx}
\usepackage{makecell}
\usepackage{amsfonts}       
\usepackage{nicefrac}       
\usepackage{microtype}      

\usepackage{wasysym}
\usepackage{pifont}

\usepackage{times}

\newcommand{\method}{\textsc{ReVPT}\xspace}

\newcommand{\ignore}[1]{}

\makeatletter
\DeclareRobustCommand\onedot{\futurelet\@let@token\@onedot}
\def\@onedot{\ifx\@let@token.\else.\null\fi\xspace}

\makeatother

\definecolor{MyBlue}{rgb}{0.46, 0.50, 0.61}
\definecolor{MyDarkBlue}{rgb}{0,0.08,0.8}
\definecolor{MyDarkGreen}{RGB}{45,155,45}
\definecolor{MyDarkRed}{rgb}{0.8,0.02,0.02}
\definecolor{MyOrange}{rgb}{1.0, 0.4, 0.2}
\definecolor{MyPurple}{RGB}{111,0,255}
\definecolor{MyRed}{rgb}{0.8,0.0,0.0}
\definecolor{MyGold}{rgb}{0.75,0.6,0.12}
\definecolor{MyDarkgray}{rgb}{0.66, 0.66, 0.66}
\definecolor{MyBrown}{rgb}{0.65, 0.16, 0.16}
\definecolor{MyMutedRose}{rgb}{0.58, 0.29, 0.35}
\definecolor{JiayuanColor}{rgb}{0.60,0.43,0.48}
\definecolor{erranColor}{rgb}{24, 40, 113}

\definecolor{citecolor}{HTML}{696FAD}

\definecolor{boxbackground}{HTML}{F0F7FF}  
\definecolor{boxborder}{HTML}{D0D9E5}      
\definecolor{accentblue}{HTML}{4A86E8}     
\definecolor{lightblue}{HTML}{EEF3FF}  
\definecolor{bordergray}{HTML}{CCCCCC}  
\definecolor{headerblue}{HTML}{2C5AA0}  

\definecolor{lavenderframe}{HTML}{E6E6FA}  
\definecolor{lighterlav}{HTML}{F5F5FF}  
\definecolor{codegray}{rgb}{0.5,0.5,0.5}  
\definecolor{codepurple}{HTML}{483D8B}  
\definecolor{backcolour}{HTML}{F5F5FF}  

\newtcolorbox{promptbox}[2][]{ 
    enhanced,
    breakable,
    boxsep=5pt,
    left=9pt,
    right=7pt,
    top=5pt,
    bottom=5pt,
    colback=boxbackground,
    colframe=boxborder,
    boxrule=0.5pt,
    arc=4pt,
    frame hidden, 
    borderline west={3pt}{0pt}{accentblue},
    shadow={0.5pt}{0.5pt}{1.5pt}{black!10},
    fontupper=\normalsize,
    title=#2, 
    colbacktitle=accentblue, 
    coltitle=white,         
    fonttitle={\fontsize{9}{11}\selectfont\bfseries}, 
    attach boxed title to top left={yshift=-2.5mm, xshift=3.2mm},
    boxed title style={
        enhanced,
        left=3pt,
        right=3pt,
        top=1pt,    
        bottom=1pt, 
        boxsep=2pt,
        arc=3pt,
        boxrule=0pt,
        colback=accentblue,
    },
    #1 
}

\newtcolorbox{notitlepromptbox}[1][]{
    enhanced,
    breakable,
    boxsep=5pt,          
    left=9pt,            
    right=7pt,           
    top=5pt,             
    bottom=5pt,          
    colback=boxbackground,
    colframe=boxborder,
    boxrule=0.5pt,
    arc=4pt,             
    frame hidden,
    borderline west={3pt}{0pt}{accentblue},  
    shadow={0.5pt}{0.5pt}{1.5pt}{black!10},  
    notitle,
    fontupper=\normalsize,    
    #1
}

\newtcolorbox{onebox}[2][]{
    enhanced, 
    center title,
    left*=0pt, right*=0pt,
    boxsep=2pt, left=5pt, right=5pt,
    skin first=enhanced,
    skin middle=enhanced,
    skin last=enhanced,
    colframe = mybluetitle!90,
  colback  = mybluetitle!10,
    fonttitle=\bfseries\rmfamily\fontfamily{phv}\selectfont,
    title={\footnotesize\strut{#2}  \refstepcounter{subsubsection} \addcontentsline{toc}{subsubsection}{\string\numberline{\thesubsubsection}#2}
    },
    #1
}

\newcommand{\header}[1]{\smallskip \noindent\textbf{{#1}}\xspace}

\title{Reinforced Visual Perception with Tools}

\runningtitle{Technical Report: Reinforced Visual Perception with Tools}

\author[1]{Zetong Zhou}
\author[1,3]{Dongping Chen\textsuperscript{$\ddagger$}}
\author[2]{Zixian Ma}
\author[1]{Zhihan Hu}
\author[1]{Mingyang Fu}
\author[1]{Sinan Wang}

\author[1]{\\Yao Wan\dag}
\author[4]{Zhou Zhao}
\author[2]{Ranjay Krishna\dag}

\affil[1]{ONE Lab, HUST}
\affil[2]{University of Washington}
\affil[3]{University of Maryland}
\affil[4]{Zhejiang University}

\correspondingauthor{Yao Wan: wanyao@hust.edu.cn, Ranjay Krishna: ranjay@cs.washington.edu, Dongping Chen: dongping@umd.edu.}

\begin{document}

\begin{abstract}
Visual reasoning, a cornerstone of human intelligence, encompasses complex perceptual and logical processes essential for solving diverse visual problems. While advances in computer vision have produced powerful models for various perceptual tasks, leveraging these for general visual reasoning remains challenging. Prior work demonstrates that augmenting LLMs with vision models via supervised finetuning improves performance, but faces key limitations such as expensive data generation, reliance on careful data filtering, and poor generalization. To address these issues, we propose \method to enhance multi-modal LLMs’ abilities to reason about and use visual tools through reinforcement learning. We introduce a novel RL algorithm based on GRPO, designed to train models to reason with a suite of four visual tools. Through extensive experiments, we show that our method achieves state-of-the-art performance on several perception-heavy benchmarks, including SAT, CV-Bench, BLINK and MMStar, outperforming the supervised and text-based RL finetuning baselines. Notably, Our \method-3B and \method-7B outperform the instruct models by 9.03\% and 9.44\% on CV-Bench. Finally, we bring to the community new insights on RL-based visual tool-usage through extensive ablations. Our code is available at \href{https://github.com/ls-kelvin/REVPT}{https://github.com/ls-kelvin/REVPT}.
\end{abstract}

\maketitle

\begin{figure}[!b]
    \centering
    \begin{minipage}[b]{0.45\linewidth}
        \centering
        \includegraphics[width=\textwidth]{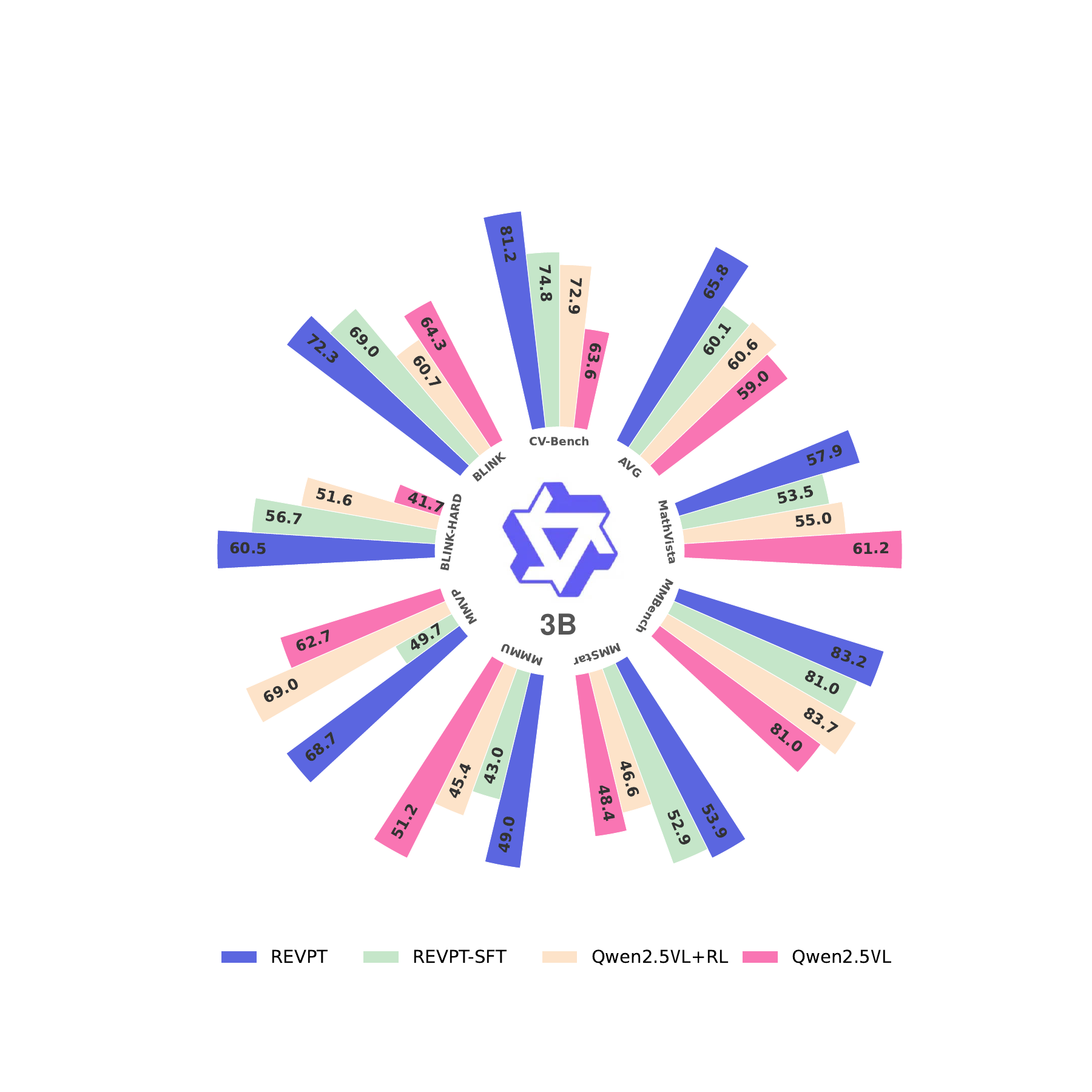}
        \label{fig:static-3b}
    \end{minipage}
    \hspace{0.02\linewidth}
    \begin{minipage}[b]{0.43\linewidth}
        \centering
        \includegraphics[width=\textwidth]{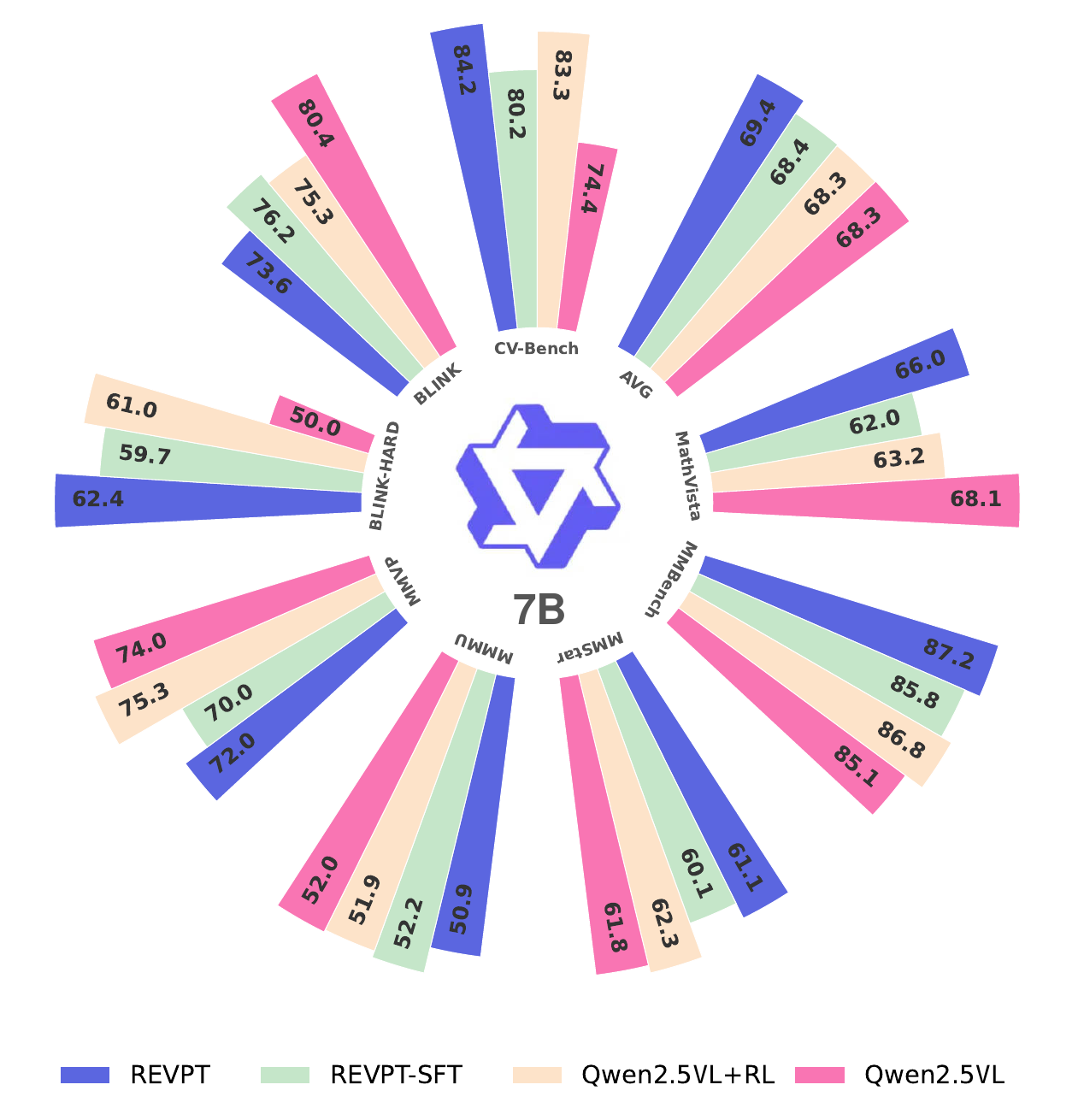}
        \label{fig:static-7b}
    \end{minipage}
    \caption{Our \method-3B and 7B models outperform their instruct and text-only GRPO counterparts on perception-centric tasks while maintaining strong general capabilities across multiple benchmarks.}
    \label{fig:performance}
\end{figure}

\section{Introduction}
\label{sec:intro}
Visual reasoning is a core component of human intelligence. It enables us to solve a wide range of problems, from daily activities such as navigation to challenging geometry problems. Unlike verbal reasoning, visual reasoning is a more complex multifaceted process that requires not just straightforward logical reasoning but also sound visual perception, which further relies on atomic abilities such as object recognition, edge detection, depth estimation, \emph{etc.} Due to the complexity of visual perception, the computer vision community has developed specialized models for different perceptual tasks, such as RecognizeAnything for object recognition, DepthAnything for depth estimation, and SegmentAnything for segmentation~\citep{depth_anything_v2, liu2023grounding, zhang2023recognize}. 


\begin{figure*}[!t]
    \centering
    \includegraphics[width=\textwidth]{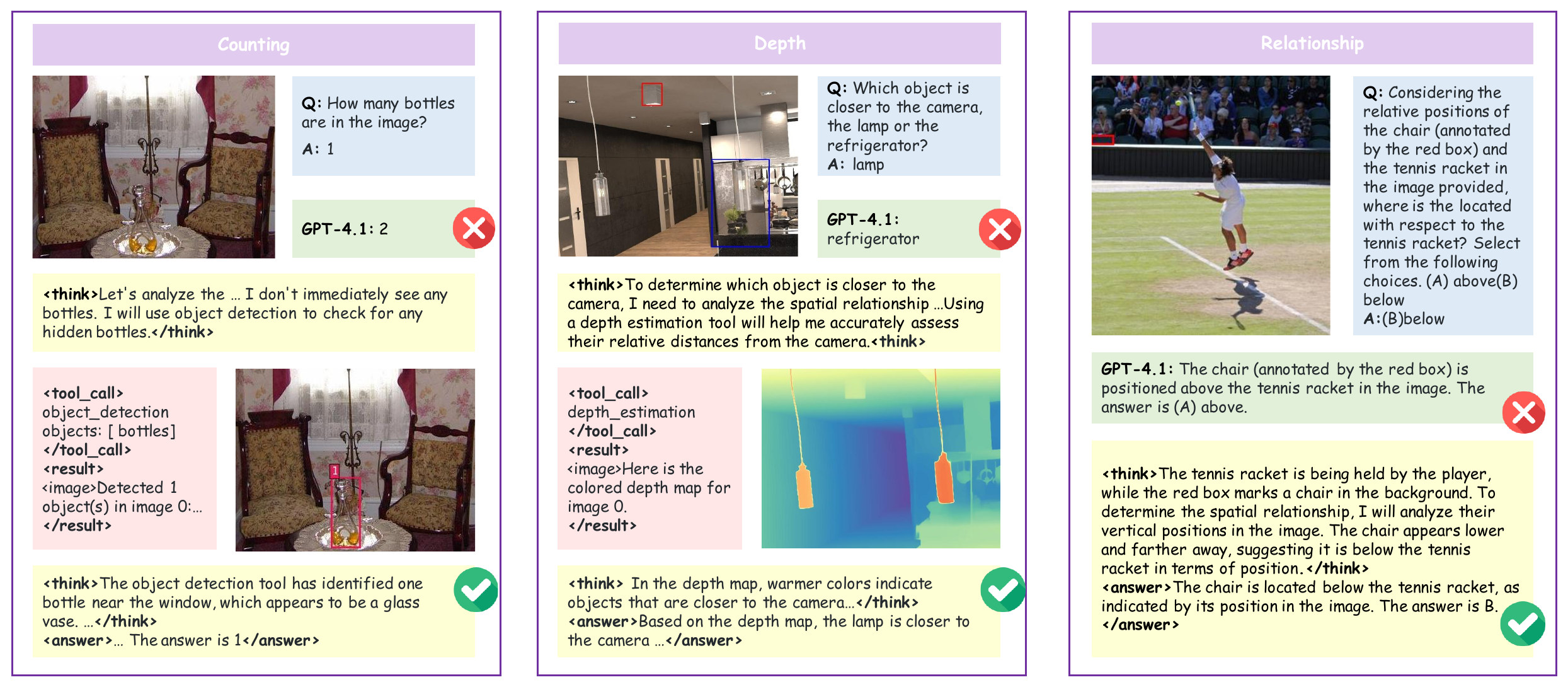}
    \caption{\method-3B and 7B are able to solve problems through thinking and selectively employing visual tools to reach strong visual perception. We illustrate some  step-by-step visual reasoning breakdowns for challenging examples that model leveraging visual tools to solve complex perception tasks and outperform commercial models GPT-4.1.}
    \label{fig:example}
\end{figure*}

To leverage these advances in perception for enhanced visual reasoning and problem-solving, prior work attempts to augment (multimodal) language models with vision specialists.  For example, VisProg first showcases that augmenting large language models (LLMs) with various vision models boosts models' performance on diverse visual tasks~\citep{Gupta2022VisProg}. Similarly, VisualSketchPad finds that prompting GPT-4o to use sketching tools and depth models improves its performance on challenging perception and visual math benchmarks by large margins~\citep{hu2024visualsketchpadsketchingvisual}. More recently, a few works demonstrate that open-source multimodal language models also benefit from using visual tools and reasoning about their outputs after supervised finetuning on tool-use data~\citep{qi2024cogcomtrainlargevisionlanguage, liu2023llavapluslearningusetools}.

Nonetheless, there are a few limitations to learning to reason with visual tools with supervised finetuning (SFT): first, it relies on expensive data curation. Prior work heavily relies on expensive commercial models like GPT-4 to generate high-quality tool-use and reasoning traces~\citep{qi2024cogcomtrainlargevisionlanguage, ma2024taco, shao2024visualcot}. Second, it often requires careful data filtering. Previous efforts reveal that aggressive filtering based on data format, answer correctness and other heuristics is crucial to performance gains~\citep{hu2024visual, ma2024taco}. Most importantly, supervised finetuning results in limited generalization: it relies on offline trajectories that pre-define particular tools to use; there is no incentive for the model to explore alternative tools or adapt to new environment outputs, thus limiting model's generalization to unseen tools or problems~\citep{guo2025deepseek, chen2025sft}.

To mitigate these limitations, we borrow inspiration from recent work on enhancing LLM reasoning and generalization with reinforcement learning (RL)~\citep{zhang2025nemotron, guo2025deepseek, huang2025vision, zhan2025vision, shen2025vlm} and propose \textbf{Re}inforced \textbf{V}isual \textbf{P}erception with \textbf{T}ools (\method).
\method enhances multimodal language model's visual perception and general question-answering by training them to reason with visual tools via RL. \method consists of cold-start and GRPO stages to enable effective RL on multimodal language models with 4 visual tools. Unlike SFT, where all the tool-use reasoning traces need to be generated and filtered in advance, we only need to select the appropriate visual questions to perform RL on. Moreover, instead of supervising the model with one correct tool-use trajectory for each question, RL allows the model to explore and learn from different possible solutions and incentivizes the model to output the best one. 

Our experiments with models of different sizes---3B and 7B---demonstrate that \method achieves state-of-the-art performance (Figure \ref{fig:performance}), consistently outperforming SFT baselines across five perception-intensive benchmarks, including MMVP, MMStar, CV-Bench, BLINK, and BLINK-Hard. Notably, our approach surpasses the original instruct-tuned models by significant margins, achieving over 5\% on average across all benchmarks, with a significant 9\% improvement on CV-Bench. Furthermore, our models outperform commercial models GPT-4.1 and Gemini-2.0-Flash on the challenging BLINK-Depth and Relation subsets. Lastly, we contribute a valuable resource to the community by fully open-sourcing \method along with its code and datasets.  

\section{\method: Thinking with Images and Tools}
\label{sec:method}

\begin{figure*}[!t]
    \centering
    \includegraphics[width=0.9\textwidth]{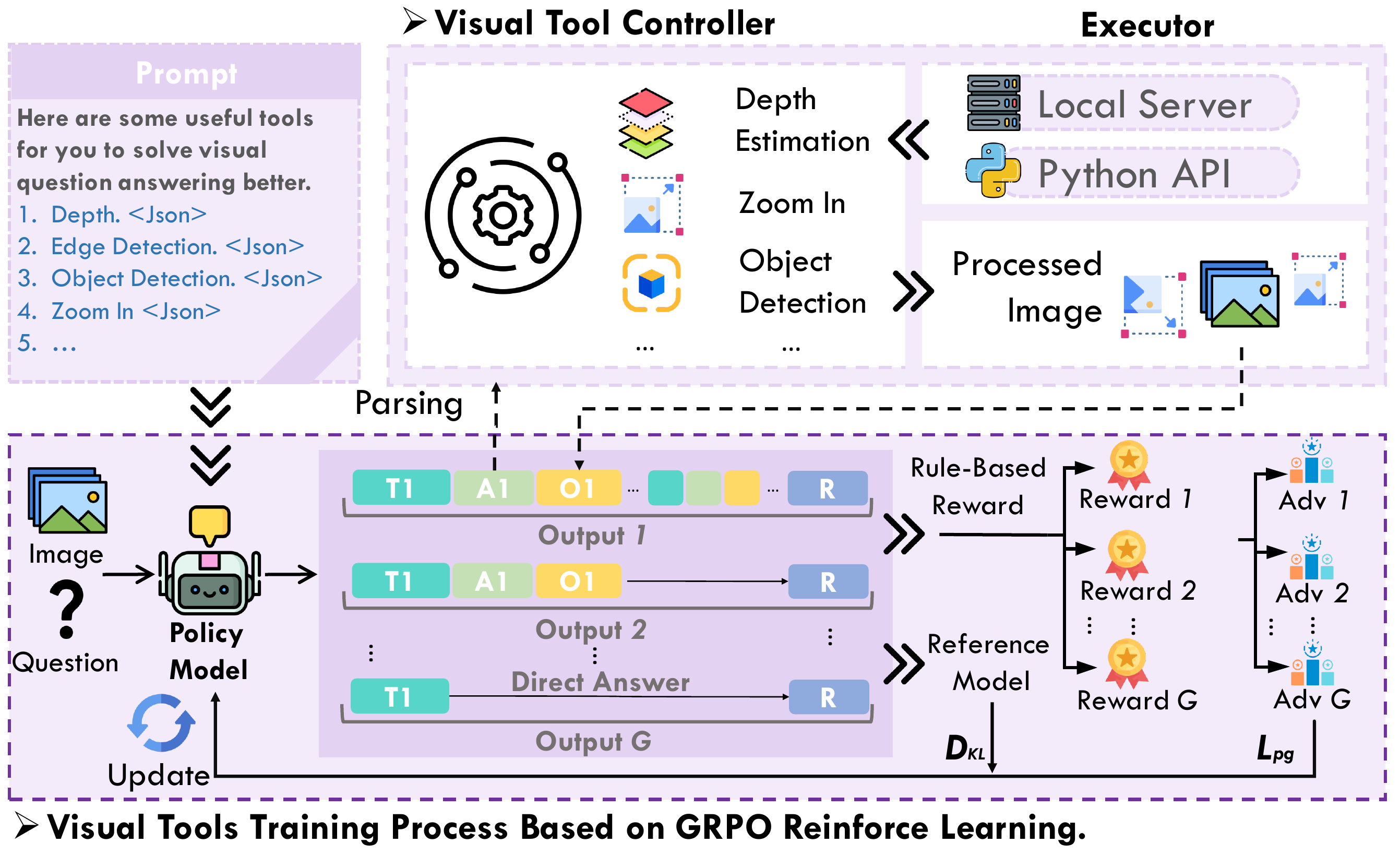}
    \caption{An overall pipeline of our \method. \textbf{(Top)}: Model-generated tool requests are managed by a local environment-based Tool Controller, which independently deploys visual tool services (\emph{e.g.}, Depth, Object Detection). These tools' outputs are then fed back to the LVLM for iterative reasoning. \textbf{(Bottom)}: When processing a visual reasoning problem, \method employs K-turn rollouts where the model interacts with the tool environment to learn an adaptive policy, culminating in the final model.}
    \label{fig:overview}
\end{figure*}

We introduce \method, a novel two-stage framework designed to train multi-modal language models for better visual perception. It is built upon the GRPO algorithm \citep{shao2024deepseekmath}, aiming to enhance the model's ability to tackle complex visual problems by integrating visual processing tools as reasoning steps. 
Given a multi-modal query, \method agent first generates a reasoning process about analyzing the query and deciding whether to call tools or answer directly. Then, by iteratively analyzing the results from the tools, the model generates a final response to the query. The overall architecture are illustrated in Figure~\ref{fig:overview}.


\subsection{Preliminary: GRPO}

Group Relative Policy Optimization (GRPO) is a reinforcement learning algorithm that evaluates policy performance by directly comparing a group of candidate responses.

The process begins with a prompt $q$, for which the policy $\pi_\theta$ samples a set of $N$ responses, $\{o_1, o_2, \dots, o_N\}$. Each response is evaluated by a reward function, yielding rewards $r_i = R(q, o_i)$. GRPO then calculates a group-relative advantage $A_i$ for each response by normalizing its reward against the statistics of the entire group:
\begin{equation}
A_i = \frac{r_i - \operatorname{mean}\{r_1, r_2, \dots, r_N\}}{\operatorname{std}\{r_1, r_2, \dots, r_N\}}\,.
\end{equation}
The policy is then updated to responses with higher relative advantages by maximizing a clipped surrogate objective function. This objective is designed to ensure stable training by preventing excessively large policy updates. The full objective is:
\begin{equation}
\begin{aligned}
J_{\mathrm{GRPO}}(\theta) 
&= \mathbb{E}_{q \sim P(Q),\{o_i\}_{i=1}^G \sim \pi_{\theta_{\mathrm{old}}}(O|q)} 
\Biggl[
  \frac{1}{G} \sum_{i=1}^G
  \min\!\Bigl(s_iA_i,\,
    \mathrm{clip}\!\Bigl(
      s_i,
      1-\varepsilon,\,
      1+\varepsilon
    \Bigr)
    A_i
  \Bigr)  -\,\beta\,\mathbb{D}_{\mathrm{KL}}\bigl(\pi_{\theta}\,\big\|\,\pi_{\mathrm{ref}}\bigr)
\Biggr]\,.
\end{aligned}
\label{eq1}
\end{equation}
Here, $s_i = \frac{\pi_\theta(o_i|q)}{\pi_{\theta_{old}}(o_i|q)}$ is the importance sampling ratio that measures the change between the new policy $\pi_\theta$ and the old policy $\pi_{\theta_{old}}$ used to generate the samples. The Kullback-Leibler (KL) divergence penalty, $\mathbb{D}_{KL}(\pi_\theta || \pi_{\text{ref}})$ regularizes the policy update by penalizing large deviations from a reference policy $\pi_{\text{ref}}$.

\subsection{Visual Tools}

The efficacy of leveraging visual tools for visual reasoning has been well-established. We choose to focus on four specific visual tools as they are the most suitable for perception-oriented tasks. While we experiment with a larger suite of tools, we find that the model is reluctant to use them. Our framework integrates these four high-performing visual tools, enabling their execution and subsequent result incorporation during the MLLM inference process to construct a comprehensive rollout. While Table~\ref{tab:vision-tools} offers a detailed summary of each tool's parameters and specifications, their core functionalities and representative use cases are elaborated upon below.

    \noindent$\triangleright$ 
    \textbf{\textsc{Object Detection}}~\citep{fu2025llmdet} is capable of open-vocabulary detection. It takes an input image $I_{in}$ and a textual query $q_{text}$ to locate instances of described objects, submitting their boundaries $\mathcal{B}_{out}$ for objects localization. 
    

    \noindent$\triangleright$ 
    \textbf{\textsc{Zoom In}} facilitates focused analysis by taking an input image $\mathcal{I}_{in}$  and a specified region of interest  $\mathcal{R}$ to output a magnified view  $\mathcal{I}_{focus}$ of that region. It helps model to concentrate on pertinent areas, thereby mitigating the influence of irrelevant information and amplifying core features.

    \noindent$\triangleright$ 
    \textbf{\textsc{Edge Detection}} is designed to identify and delineate the perimeters of objects and significant textural variations, this tool transforms an input image $\mathcal{I}_{in}$ into a feature map $\mathcal{E}_{out}$ that emphasizes these structural edges. This grants models a sharper perception of object silhouettes and surface patterns.

    \noindent$\triangleright$ 
    \textbf{\textsc{Depth Estimation}}~\citep{depth_anything_v2} computes spatial depth from a monocular visual input. Given an image $\mathcal{I}_{in}$ , it generates a corresponding depth map  $\mathcal{D}_{out}$ which encodes the estimated distance of scene elements from the observer. We use \textsc{depth\_anything\_v2} to serve as the tool, equipping models with a more nuanced understanding of three-dimensional arrangements and the relative positioning of objects.

\begin{table*}[!t]
\centering
\resizebox{\textwidth}{!}{%
\begin{tabular}{llll}
\toprule
\textbf{Tool}                    & \textbf{Description}              & \textbf{Arguments}               & \textbf{Tool Output}           \\
\midrule
\textsc{Object Detection} & Open-vocabulary object detection & Image + Text Prompt & Annotated Image + Boxes \\
\textsc{Zoom In} & Crop a region and augment it & Image + Coordinates + Factor & Cropped Image \\
\textsc{Edge Detection} & Edge detection using scharr algorithm & Image & Edge Map \\
\textsc{Depth Estimation} & Return depth map in colorful style & Image & Depth Map \\
\bottomrule
\end{tabular}}
\caption{Visual tools integrated within \method. We illustrate their arguments, result formats, and  core funtions description.}
\label{tab:vision-tools}
\end{table*}

\begin{figure}[!ht]
    \centering
    \includegraphics[width=0.95\linewidth]{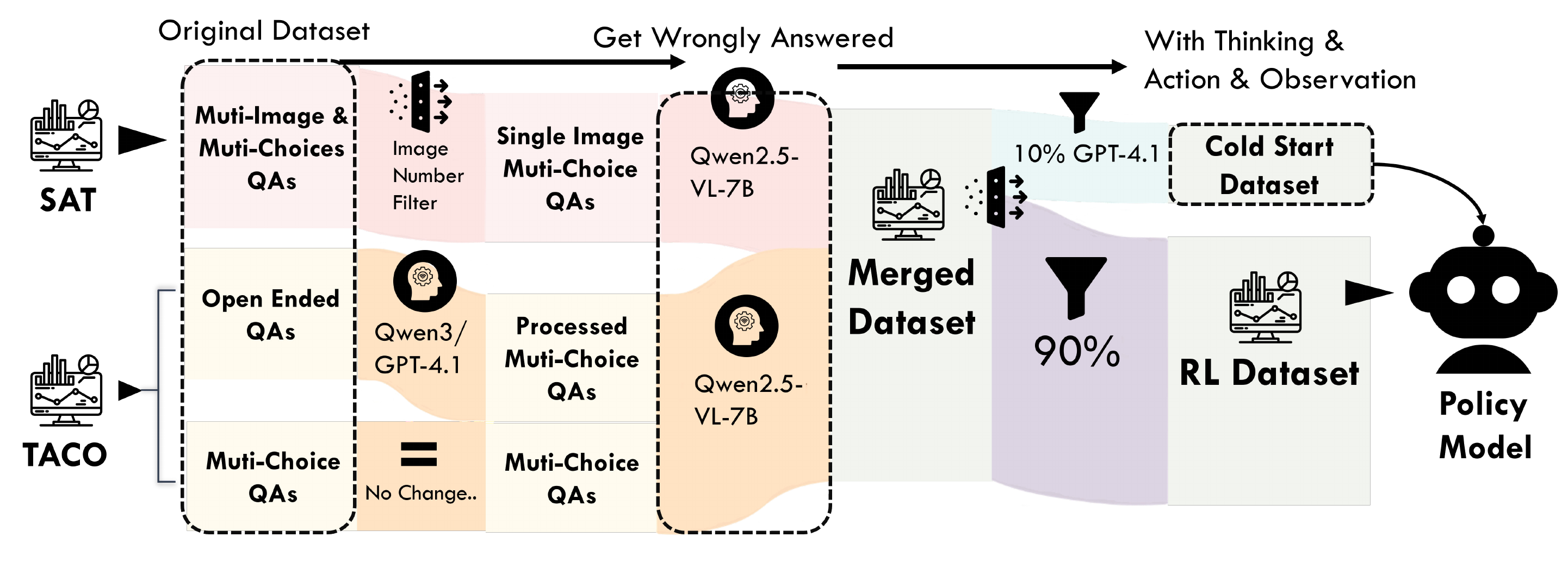}
    \caption{Reinforced visual tool-usage training requires high-quality and verified data. We transform two datasets into multiple-choice question, then filter out too easy or hard questions for Qwen2.5-VL-7B by multiple inference.}
    \label{fig:dataflow}
\end{figure}

\subsection{Cold Start}
Initially, our objective is to train a multimodal agent using the R1-Zero method. However, during the training process, we observe a progressive decline in the agent's propensity to utilize tools. This phenomenon likely stemmed from the fact that solving visual tasks did not inherently require tool usage, and reasoning based on processed images represented a distribution shift from the model's initial training data. Consequently, we opt to incorporate cold-start data during the initial training phases as shown in Figure \ref{fig:dataflow}. This strategy aims to facilitate the model's initial acquisition of tool utilization skills for solving visual problems.

Currently, we employ GPT-4.1 as the agent to synthesize high-quality data for tool-augmented reasoning in visual problem solving. We require the GPT model to follow a predefined format when answering questions. Specifically, it should first engage in a reasoning process, considering the potential utility of external tools, and then employ a multi-turn generation strategy to address the problem. We explicitly encourage the model in our prompt to generate more tool-assisted rollouts. Subsequently, we filter out rollout trajectories generated by the GPT model that contain errors.

After synthesizing tool-integrated reasoning data, we perform supervised fine-tuning on it. Denote the query as $Q^{i}$, the rollout trajectory $\tau^i$ as a sequence of actions $a_t^i$ and observations $o_t^i$. We optimize the cross-entropy loss: 
\begin{align}
\mathcal{L}_{\mathrm{SFT}}(\theta) 
&=-\frac{1}{N}\sum_{i=1}^{N}
\log P_{\theta}\bigl(\tau^{i} \mid Q^{i}\bigr) \tag{6}\\
&= -\frac{1}{N}\sum_{i=1}^{N}\sum_{t=1}^{n^{i}}
\log P_{\theta}\bigl(a_{t}^{i},\mid Q^{i},\,a_{<t}^{i},\,o_{<t}^{i}\bigr)\,. \tag{7}
\end{align}
By minimizing $\mathcal{L}_{\mathrm{SFT}}$, the model acquires a robust Cold‐Start policy for sequential vision‐tool invocation, providing a solid foundation for the subsequent reinforcement learning phase. 

\subsection{Reward Design}
Vision tasks often possess ground truth data, facilitating the use of verified rewards to evaluate model's responses. This approach circumvents the introduction of neural network-based reward models, thereby preventing reward hacking.
\begin{itemize}[leftmargin=*,itemsep=0pt]
    \item \textbf{Correctness.} We restructure the dataset into a verifiable format. The model is required to put the final direct answer or option in the box, enabling reliable rule-based verification of correctness. If the answer match the ground truth, it gets through checking.
    \item \textbf{Format.} In each turn, the response should enclose its thinking process in \texttt{<think></think>} and enclose its tool call in \texttt{<tool\_call></tool\_call>} or answer in \texttt{<answer></answer>}. If the response matches the pattern, it gets through format checking. Given the query $q$ and the rollout $o$, the reward is defined as:
\[
Reward(q,o) = 
\begin{cases}
1, & If \ FormatCorrect(o) \\ & \quad \land  \ AnswerCorrect(o)  \\ 
-1, & otherwise
\end{cases}
\]
\end{itemize}

We adopt this naive binary reward design because previous research has demonstrated the effectiveness of using simple, rule-based reinforcement learning to produce strong reasoning behaviors \citep{zhou2025r1zerosahamomentvisual}.

\section{Experiment}
\label{sec:experiment}

\subsection{Experiment Setups}

\header{Models.} We conduct most of our experiments on Qwen2.5-VL-3B-Instruct and Qwen2.5-VL-7B-Instruct \citep{Qwen2VL} for their strong visual perception capability and how different \textit{``perception baseline''} contribute to final capability.

\header{Dataset Construction.} A high quality cold-start dataset and a difficult dataset for RL training is very important for  model to learn test-time scaling itself \citep{yu2025dapo}. Therefore, we filter the SAT dataset \citep{ray2024sat} and Taco \citep{ma2024taco} training set with Qwen2.5-VL-7B-Instruct and retain the samples it answers incorrectly. 
SAT is a synthetic VQA dataset designed to improve a model's visual perception, which has been shown to be effective in prior work \citep{zhou2025r1zerosahamomentvisual}. TACO dataset was also utilized in previous research \citep{ma2024taco} for multi-modal chain-of-thought SFT. We selected it based on the assessment that its VQA content is better suited for tool-assisted responses.
Then, we split it into cold-start data synthetic and training data. To construct high-quality cold-start dataset, we leverage GPT-4.1 to synthesize tool trajectory and retain the subset that it answers correctly. Finally, we obtain a 1.5k cold-start dataset with well-curated reasoning chain and tool trajectory and 20k question for RL process. 

\begin{table*}[!t]
\centering
\setlength{\tabcolsep}{3pt}
\resizebox{\textwidth}{!}{
\begin{tabular}{l|c|ccccc|c|c|c|c|c|c|c}
\toprule[1.5pt]
\multirow{2}{*}{\textbf{Model}} & \multirow{2}{*}{\multirow{2}{*}{\textbf{Avg.}}} & \multicolumn{5}{c|}{\textbf{CV-Bench}} & \multicolumn{1}{c|}{\multirow{2}{*}{\textbf{BLINK(sub)}}} & \multicolumn{1}{c|}{\multirow{2}{*}{\textbf{MMVP}}} & \multicolumn{1}{c|}{\multirow{2}{*}{\textbf{MMSTAR}}} & \multicolumn{1}{c|}{\multirow{2}{*}{\textbf{BLINK-HARD}}} & \multicolumn{1}{c|}{\multirow{2}{*}{\textbf{MMMU}}} & \multicolumn{1}{c}{\multirow{2}{*}{\textbf{MMBench}}} & \multicolumn{1}{c|}{\multirow{2}{*}{\textbf{MathVista}}}\\
\cmidrule(lr){3-7} 
 & & \textbf{Count} & \textbf{Relation} & \textbf{Depth} & \textbf{Distance} & \textbf{Avg.} & & & & & & &\\
\midrule
\rowcolor{gray!10} Gemini-2.0-Flash & 76.09 & 71.95  & 86.92  & 87.50  & 82.17  & 81.50  & 76.37 & 79.34  & 69.40 & 68.28 & 71.7 & 89.05 & 73.1\\
\rowcolor{gray!10} GPT-4.1         & 76.06 & 67.77  & 92.00  & 94.50  & 89.50  & 84.76  & 68.80  & 88.00  & 69.80 & 66.13 & 74.00 & 86.60 & 70.40\\
\rowcolor{gray!10} TACO (Qwen2-VL-7B) & 58.55 & 63.32 & 81.08 & 59.83 & 57.26 & 66.00 & 65.50 & 67.00 & 49.53 & 33.13 & 44.0 & 81.80 & 41.9\\
\midrule
Qwen2.5-VL-3B-Instruct   & 58.98 & 68.65  & 74.92  & 76.00  & 71.67  & 72.55  & 64.34  & 62.67  & 53.40 & 41.67 & 49.00 & 80.97 & 61.20\\
Qwen2.5-VL-3B-SAT-SFT & 60.99 & 60.66 & 90.15 & 87.33 & \underline{84.50} & 79.42 & 63.79 & 58.33 & 50.84 & 56.71 & 42.11 & 80.09 & 56.66\\
Qwen2.5-VL-3B-GRPO & 60.60 & 69.16 & 81.23 & 80.17 & 61.50 & 72.90 & 60.66 & 69.00 & 46.60 & 51.61 & 45.44 & 83.69 & 54.95\\

\textbf{\method-3B (Ours)}       & 65.67 & 70.43 & \underline{92.62} & \underline{87.50} & 78.33 & \underline{81.20} & 72.35 & 68.70 & 53.87 & 60.48 & 49.00 & 83.21 & 57.90\\ 
\rowcolor{pink!20} $\Delta$ & \textcolor{red}{+6.69} & \textcolor{red}{+1.78} & \textcolor{red}{+17.70} & \textcolor{red}{+11.50} & \textcolor{red}{+6.66} & \textcolor{red}{+8.65} & \textcolor{red}{+8.06} & \textcolor{red}{+6.03} & \textcolor{green}{-0.47} & \textcolor{red}{+18.81} & 0.00 & \textcolor{red}{+2.24} & \textcolor{green}{-4.20}\\ \midrule
Qwen2.5-VL-7B-Instruct & 68.23 & 64.97 & 88.46 & 72.00 & 74.00 & 74.41 & \textbf{80.41} & \underline{74.00} & \underline{61.80} & 50.00 & \textbf{52.00} & 85.08 & \textbf{68.10}\\
Qwen2.5-VL-7B-SAT-SFT & 62.70 & 60.28 & 84.46 & 85.33 & 82.67 & 77.03 & 67.91 & 62.67 & 49.80 & 57.00 & 45.22 & 82.49 & 59.50\\
Qwen2.5-VL-7B-GRPO & \underline{69.38} & \underline{72.59} & \textbf{95.08} & 83.17 & \textbf{84.83} & 83.32 & \underline{75.87} & \textbf{75.33} & \textbf{62.27} & \underline{61.00} & \underline{51.88} & \underline{86.83} & 63.20\\
\textbf{\method-7B (Ours)} & \textbf{69.42} & \textbf{74.11} & 92.31 & \textbf{88.67} & 82.67 & \textbf{84.23} & 73.64 & 72.00 & 61.07 & \textbf{62.37} & 50.89 & \textbf{87.20} & \underline{66.00} \\
\rowcolor{pink!20} $\Delta$ & \textcolor{red}{+5.73} & \textcolor{red}{+9.14} & \textcolor{red}{+3.85} & \textcolor{red}{+16.67} & \textcolor{red}{+8.67} & \textcolor{red}{+9.82} & \textcolor{green}{-6.77} & \textcolor{green}{-2.00} & \textcolor{green}{-0.73} & \textcolor{red}{+12.37} & \textcolor{green}{-1.11} & \textcolor{red}{+2.12} & \textcolor{green}{-2.10}\\ 
\bottomrule[1.5pt]
\end{tabular}
}
\caption{Comparison between different models and method on vision-centric benchmarks. Our \method outperform direct SFT and text-space GRPO in perception-specific tasks and preserve general capability.}
\label{tab:result}
\end{table*}

\header{Baselines.} We compare \method against the following models and approaches: \textbf{\textit{(1)} Commercial Models}: We select GPT-4.1 \citep{openai2025gpt41} and Gemini-2.0-Flash \citep{google2024gemini2.5flash} as strong multimodal baselines. Both are evaluated in a zero-shot setting without external tools as strong generalist benchmarks. \textbf{\textit{(2)} \textsc{Taco}}: Learns to invoke 15 external tools (\emph{e.g.}, OCR, depth, \emph{etc.}) by generating Chain-of-Thought-and-Action (CoTA) sequences via supervised learning on synthetic data. TACO typically executes tools within a single process, contrasting with our RL-based approach and distributed architecture \citep{ma2024taco}. \textbf{\textit{(3)} Qwen-Instruct}: We implement two instruct version models without any tool usage, employing basic prompt formats for a robust evaluation. \textbf{\textit{(4)} Qwen-SAT-SFT}: Two models after supervised finetuning with the SAT training set (13k samples) as a strong baseline. We also include an enhanced SFT baseline with rephrased answers generated by the model itself as reported by previous research \citep{wang2024theoretical}. \textbf{\textit{(5)} Qwen-SFT (cold start)}: The model state after cold-start training. \textbf{\textit{(6)} Text-based RL:} An RL agent trained similarly to \method but without any visual feedback from tools. This follows the native GRPO training in VisualThinker-R1-Zero \citep{zhou2025r1zerosahamomentvisual}.

\header{Evaluation.} We select 8 multi-modal benchmarks covering diverse capabilities with a focus on visual perception and reasoning tasks. This includes perception-centric benchmarks CV-Bench \citep{tong2024cambrian1}, Blink \citep{fu2024blink}, MMVP \citep{tong2024eyes}, Blink-Hard \citep{bigverdi2024perception}. We also include MMStar \citep{chen2024mmstar}, MMMU-val \citep{yue2023mmmu} and MMBench \citep{MMBench} for general capabilties assessment. Reasoning benchmark MathVista \citep{lu2024mathvista} is also included. We choose subset with only one image as input. All results are reported under average of three runs.

\header{Experiment Details.} We conduct model training using configurations with 8 $\times$ NVIDIA A800 GPUs. We leverage LLaMA-Factory \citep{zheng2024llamafactory} as SFT platform and Verl as visual tool-base RL platform \citep{sheng2024hybridflow}. The training process involved two phases: \textbf{\textit{(1)} Cold-start Period}:
Models are trained for 2 epochs with a learning rate of 1e-5 and a global batch size of 64. \textbf{\textit{(2)} RL Period}: For this phase, models are trained for 200 steps to reach reward converge. We provide training loss and reward curve in Appendix.

\subsection{Experiment Results}

\header{\method outperform supervised finetuning and text-based RL on visual perception tasks.} 
As shown in Table \ref{tab:result}, our method significantly outperforms the original model on both CV-Bench and BlinkHard benchmarks, while also demonstrating improvements over SFT cold start models. The performance gains are particularly pronounced in depth and distance tasks, with improvements of 9.5\% and 13.23\% respectively compared to the base model, indicating that reinforcement learning effectively teaches the model to utilize tools and interpret tool-generated information. The consistent improvements across \method-3B (+5.81\%) and \method-7B (+5.73\%), with striking gains on perception-heavy benchmarks like CV-Bench (+8.65\% and +9.82\%), demonstrate that reinforcement learning unlocks latent capabilities inaccessible through conventional training. This ceiling elevation stems from RL's unique ability to enable exploratory tool selection and adaptive reasoning during inference, allowing models to dynamically adjust problem-solving strategies rather than being constrained to pre-defined patterns from static supervised trajectories. The fact that \method substantially outperforms not only base instruct models but also specialized baselines suggests that reward-driven optimization creates a synergistic learning dynamic that transcends single training methodologies, indicating the true potential of visual perception capabilities may be significantly higher than previously demonstrated.

\header{Performance tradeoffs in perception-specialized training.} Our training on the curated perception subset from SAT and TACO resulted in a slight reduction of the model's general performance capabilities. Direct supervised fine-tuning on the SAT dataset (SAT-SFT) revealed substantial degradation across several capabilities required for BLINK and MMStar evaluations. These findings underscore the critical relationship between cold-start sample distribution and the resulting model capabilities.
Our key insight is that developing effective visual tool-using agents requires not only sophisticated interaction environments but also carefully balanced, comprehensive, and high-quality data construction that preserves general capabilities while enhancing targeted skills.

\begin{table*}[!t]
\centering
\setlength{\tabcolsep}{2pt}
\resizebox{0.98\textwidth}{!}{
\begin{tabular}{cc|c|c|ccccc|ccc|c|c|ccc}
\toprule[1.5pt]
\multicolumn{2}{c|}{\textbf{Data Construction}} & \multirow{2}{*}{\textbf{Hint}} & \multicolumn{1}{c|}{\multirow{2}{*}{\textbf{Avg.}}} & \multicolumn{5}{c|}{\textbf{CV-Bench}} & \multicolumn{3}{c|}{\textbf{BLINK}} & \multicolumn{1}{c|}{\multirow{2}{*}{\textbf{MMVP}}} & \multicolumn{1}{c|}{\multirow{2}{*}{\textbf{MMSTAR}}} & \multicolumn{3}{c}{\textbf{BLINK-HARD}}\\
\cmidrule(lr){5-9} \cmidrule(lr){10-12}  \cmidrule(lr){15-17}
  \textsc{Tool-usage} & \textsc{General} & & & \textbf{Count} & \textbf{Relation} & \textbf{Depth} & \textbf{Distance} & \textbf{Avg.} & \textbf{Count} & \textbf{Depth} & \textbf{Relation} & & & 3 & 4 & 5 \\
\midrule
\ding{52} & \ding{52} & & 78.28 & 66.62 & 87.69 & 82.50 & 76.83 & 77.75 & 65.00 & 73.39 & \textbf{83.92} & 68.67 & 60.40 & 65.32 & 54.84 & 51.61 \\
\ding{52} & & \ding{52} & 77.26 & 65.74 & 88.62 & 82.33 & 72.83 & 76.76 & 60.00 & 79.03 & 81.12 & 69.33 & 58.33 & 61.29 & 54.03 & 59.68\\
 & \ding{52} & \ding{52} & 78.97 & \textbf{68.15} & 87.23 & 80.67 & \textbf{80.33} & 78.46 & \textbf{65.83} & 72.58 & 81.82 &69.00 & 58.33 & 63.71 & 50.81 & 41.94 \\
 \ding{52} & \ding{52} & \ding{52} & \textbf{80.86} & 66.24 & \textbf{91.54} & \textbf{87.17} & 79.17 & \textbf{80.17} & \textbf{65.83} & \textbf{79.84} & 81.82 & \textbf{72.67} & \textbf{61.80} & \textbf{68.55} &
\textbf{55.65} &
\textbf{60.48} \\
\bottomrule[1.5pt]
\end{tabular}
}
\caption{Ablation study on \method-7B SFT data construction demonstrates that synthesized tool-usage data and general data from TACO collaboratively contribute to an effective cold-start process, preserving the model's general capabilities while enabling visual tool usage. Specifically, training with synthesized data where the model provides 1-2 sentence reasoning about task suitability for visual tools proves beneficial for this dual objective.}
\label{tab:ablation_on_data}
\end{table*}

\begin{figure}[htbp]
    \centering
    \begin{minipage}[b]{0.53\linewidth}
        \centering
        \includegraphics[width=\textwidth]{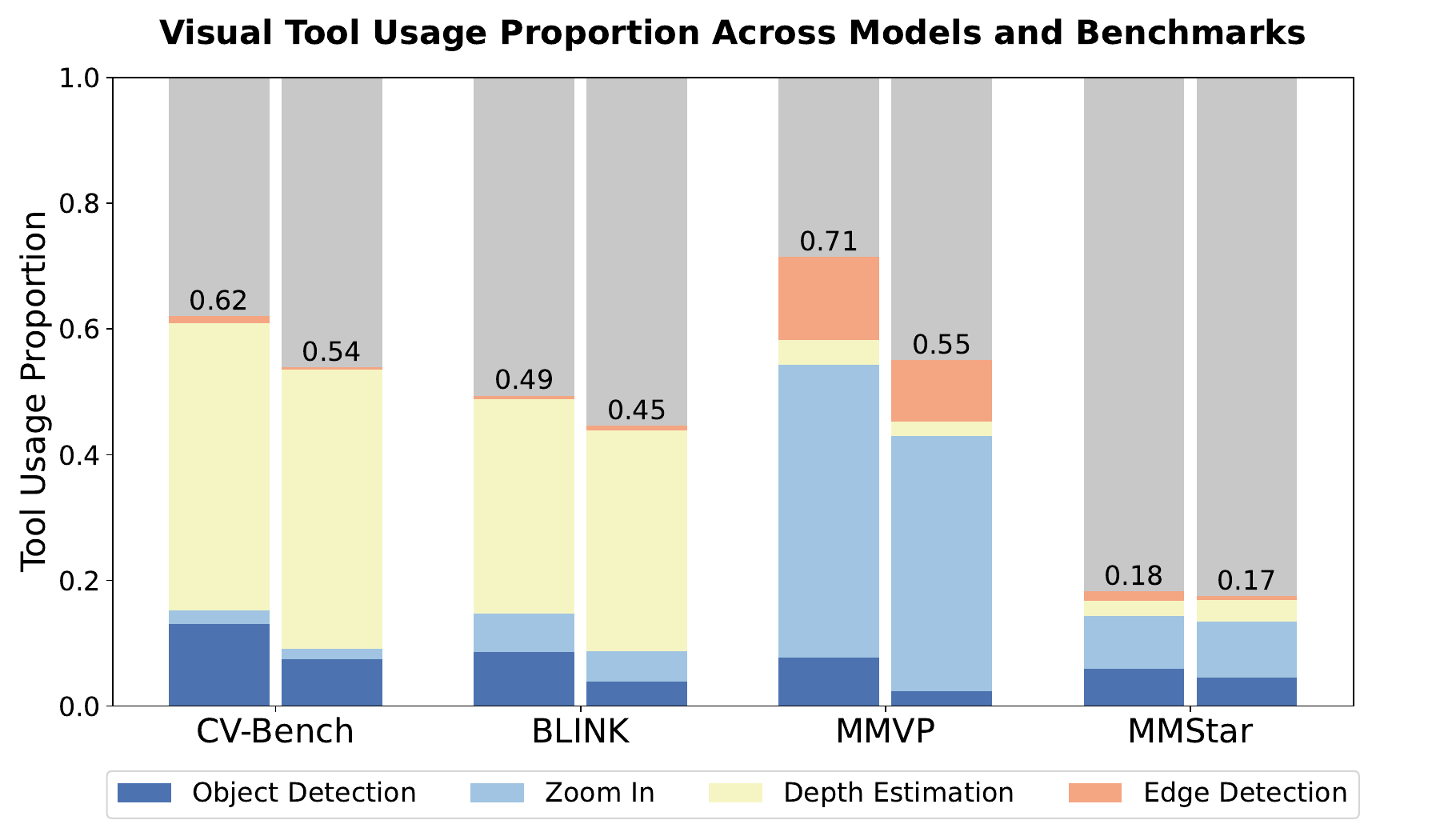}
        \caption{\method-3B effectively learns tool utilization after cold-start process. \method-3B shows bias toward object detection and depth estimation over zoom and edge detection tools due to cold-start data construction. Future work will address data balance and generalized perception objectives.}
        \label{fig:tools_propotion}
    \end{minipage}
    \hfill
    \begin{minipage}[b]{0.45\linewidth}
        \centering
        \includegraphics[width=\textwidth]{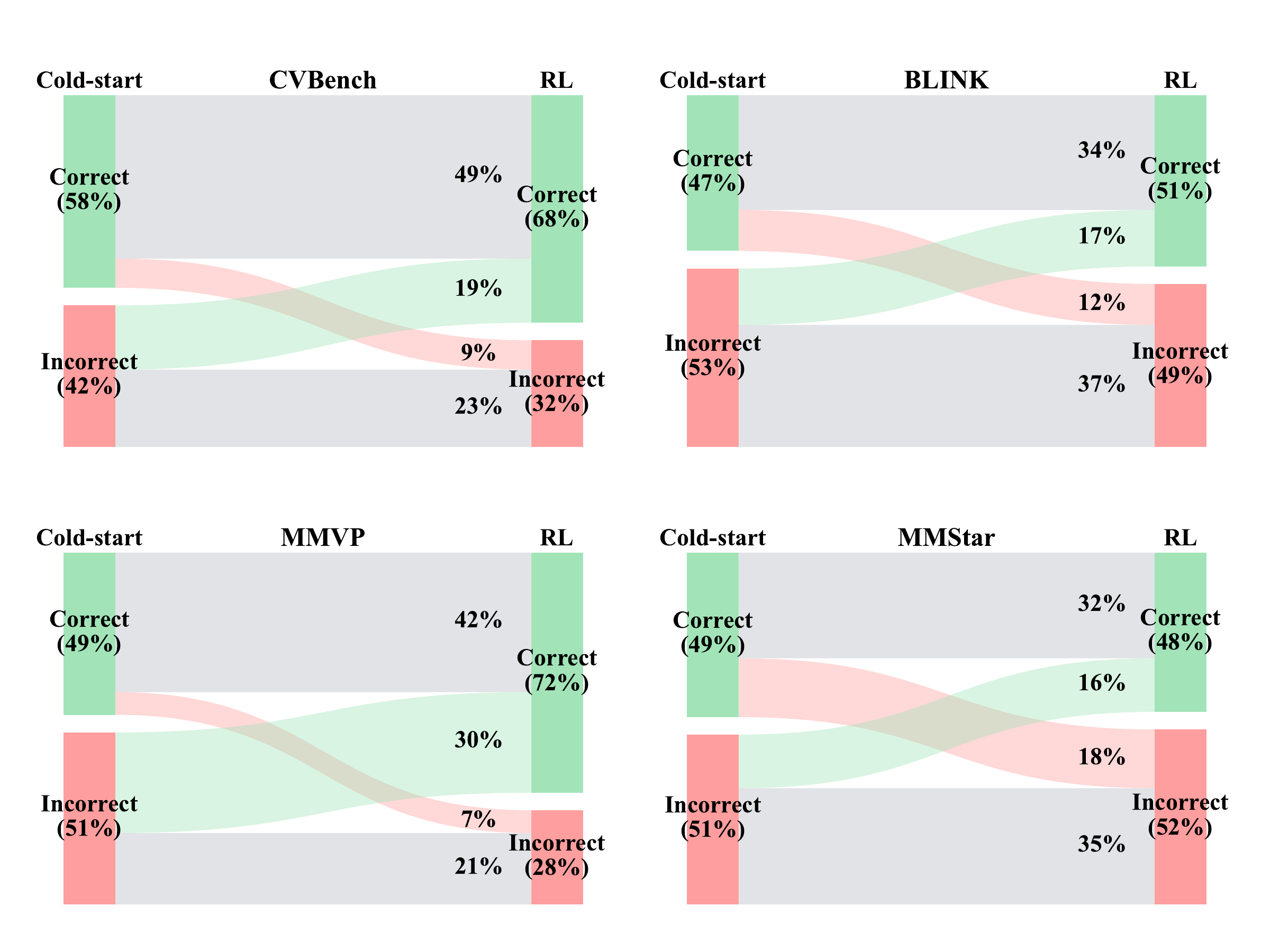}
        \caption{\method-3B learns to improve accuracy by changing its tool usage after reinforcement learning. This strategic shift results in a higher proportion of correct answers, demonstrating that the RL phase optimizes model's tool selection. }
        \label{fig:shift}
    \end{minipage}
\end{figure}

\begin{table*}[!t]
\centering
\setlength{\tabcolsep}{2pt}
\resizebox{0.98\textwidth}{!}{
\begin{tabular}{cc|c|ccccc|ccc|c|c|ccc}
\toprule[1.5pt]
\multicolumn{2}{c|}{\textbf{Visual Tools}} & \multirow{2}{*}{\textbf{RL}} & \multicolumn{5}{c|}{\textbf{CV-Bench}} & \multicolumn{3}{c|}{\textbf{BLINK}} & \multicolumn{1}{c|}{\multirow{2}{*}{\textbf{MMVP}}} & \multicolumn{1}{c|}{\multirow{2}{*}{\textbf{MMSTAR}}} & \multicolumn{3}{c}{\textbf{BLINK-HARD}}\\
\cmidrule(lr){4-8} \cmidrule(lr){9-11}  \cmidrule(lr){14-16}
 \textsc{Object} & \textsc{Depth} & &\textbf{Count} & \textbf{Relation} & \textbf{Depth} & \textbf{Distance} & \textbf{Avg.} & \textbf{Count} & \textbf{Depth} & \textbf{Relation} & & & 3 & 4 & 5 \\
\midrule
\ding{52} &  & \ding{52} & 70.05 & 91.08 & 85.17 & 77.67 & 80.40 & 60.00 & 70.16 & 76.22 & 63.67 & \textbf{54.53} & 66.13 & 57.26 & 58.06\\
 & \ding{52} & \ding{52} & 68.53 & 90.46 & 86.33 & 71.33 & 78.62 & 55.83 & 69.35 & \textbf{84.62} & 63.33 & 53.93 & 62.10 & \textbf{58.87} & 58.87 \\
\ding{52} & \ding{52} & & 66.62 & 87.54 & 86.50 & 69.67 & 76.99 & 59.17 & 73.39 & 74.83 & 51.00 & 53.47 & 61.29 & 55.65 & 54.03 \\
\ding{52} & \ding{52} & \ding{52} & \textbf{70.43} & \textbf{92.62} & \textbf{87.50} & \textbf{78.33} & \textbf{81.58} & \textbf{60.83} & \textbf{75.81} & 81.12 & \textbf{70.33} & 53.87 & \textbf{66.13} & 57.26 & \textbf{60.48}\\
\bottomrule[1.5pt]
\end{tabular}
}

\caption{Ablation study on visual tools and RL phase showing that each component within \method-3B collaboratively contribute to our outstanding results on vision perception tasks.}
\label{tab:ablation}
\end{table*}

\subsection{In-Depth Analysis}

\header{Ablation study on how to introduce tool-use capability to current models.} To address the above mentioned challenge, we conducted an ablation study on cold-start process (shown in Table \ref{tab:ablation_on_data}) to determine how to introduce tool-use agent capabilities to the current model without compromising its general performance. Our empirical experiments demonstrate that synthesizing hints about whether and why to use tools can drastically preserve the model's general capabilities. Moreover, including general data from TACO also contributes to maintaining general model capability, while incorporating tool-use data enhances the model's tool-using performance.

\header{Ablation study on visual tools.} We conduct ablation studies on the two most frequently used visual tools---object detection and depth estimation---by removing these tools from both data and instructions to validate their importance as core components of \method. The experiments follow the same cold-start and RL training procedures, with results shown in Table \ref{tab:ablation}.
When removing object detection, performance declines significantly across most metrics: BLINK relation dropped by 5 points (8.2\% decrease) from 60.83 to 55.83, and MMVP scores fall by 12.33 points (9.6\% decrease) from 70.33 to 63.33. This confirms that visual tool usage is fundamental to \method's effectiveness.

\header{Tool usage analysis.} We analyze the frequency of tool usage by cold-start and RL phase models across different benchmarks, as shown in Figure \ref{fig:tools_propotion}. On perception-intensive benchmarks like CV-Bench and MMVP, models exhibit high tool usage frequency, while on the more general MMSTAR benchmark, tool usage frequency decreases substantially.
Furthermore, we observe that object detection and edge detection usage decline significantly after RL training, resulting in an overall reduction in tool usage frequency. To figure out how vision tools contribute to model's improvement, we examine the changing portion (Figure \ref{fig:shift}) between \method-3B and base model. Across 4 benchmarks, we find that \method-3B achieves higher accuracy when using tools, indicating that visual tool-usage provide more information and select more appropriate tools for visual perception tasks.

\begin{figure*}[!t]
    \centering
    \includegraphics[width=\textwidth]{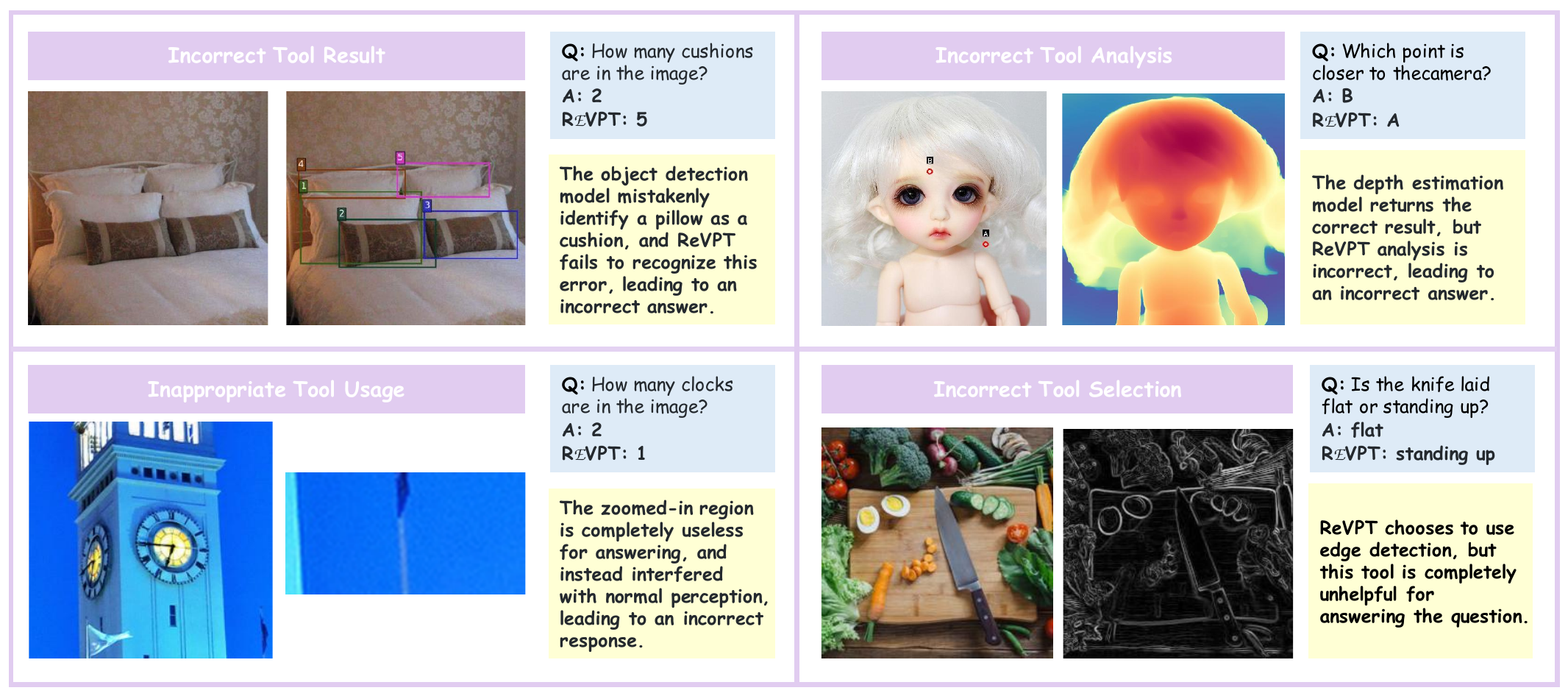}
    \caption{Case studies illustrating various failure modes of \method when using visual tools.Examples include incorrect tool output (top left) , misinterpretation of a correct tool output (top right) , inappropriate tool usage that interferes with perception (bottom left) , and the selection of an unhelpful tool for the given task (bottom right).}
    \label{fig:casestudy}
\end{figure*}

\header{Limitations of visual tools.}
During our experimental process, we observe that these visual tools often hinder rather than enhance model performance, sometimes degrading or interfering with the model's reasoning capabilities. In some cases, the model needed to correct erroneous tool outputs to arrive at accurate answers. For instance, in Figure \ref{fig:casestudy} (top left), the object detection model misclassify a mattress as a pillow, resulting in an incorrect response despite the model's inherent capability to solve the problem correctly.  Case study in Figure \ref{fig:casestudy} (top right) shows that although the depth estimation tool returned a color-coded depth map, the model do not acknowledge this information and use it during reasoning, ultimately producing an incorrect answer. We attribute this phenomenon to the model's original perception capabilities constraining its ability to effectively and accurately utilize external tools, limiting tool-use efficacy.

\section{Discussion and Future Works}
\header{Do we really need visual tool specialist?} The utility of visual tool specialists follows a non-monotonic relationship with model capability. For resource-constrained models like our 3B variants, visual tools provide a compelling shortcut to enhanced perception, with \method achieving over 9\% improvements on perception-heavy benchmarks. This aligns with findings from TACO \citep{ma2024taco}, where chain-of-thought-and-action approaches enabled performance gains of 30-50\% compared to few-shot prompting. However, as models scale to moderate sizes, the marginal benefit diminishes—larger vision encoders reduce dependency on external tools, and our analysis reveals cases where tool outputs contradict correct model intuitions, leading to degraded performance. Paradoxically, the most advanced models like GPT-o3~\citep{openai2025o3} may experience a tool renaissance through dynamic code generation, representing a shift from using pre-defined tools to creating sophisticated reasoning chains. This suggests tool integration is worthwhile across the capability spectrum, with beneficial tools evolving from compensating for perceptual deficits in small models to enabling sophisticated multi-step reasoning in large ones.

\header{Which visual tools should we select for tool-use training?} The selection of visual tools fundamentally shapes model capabilities and constrains downstream performance. Our domain-specific approach prioritized tools aligned with perceptual challenges: object detection, depth estimation, edge detection, and zoom-in functionality. While this reflects broader patterns where tool selection is guided by target applications, it reveals fundamental limitations in current approaches that constrain generalizability. The core challenge lies in the cold-start nature of tool learning, which necessitates committing to a specific tool repertoire during initial training \citep{zhang2025thyme,wang2025simple}. Our experiments initially included broader tools (\emph{e.g.}, auxiliary line drawing, highlighting, rotation) but found extremely low utilization rates, underscoring that smaller models struggle to learn multiple tools simultaneously due to limited world knowledge. The constraint is data-driven—achieving proficiency across diverse tool sets requires comprehensive training datasets. Our reliance on SAT and TACO datasets lacks the breadth for more specialized tools, suggesting that advancing visual tool learning requires systematic development of richer, more diverse training ecosystems.

\header{The Bitter Lesson of Tool Use: Human Intuitions vs. Computational Discovery.} A fundamental tension exists in tool-using AI systems: while any approach producing correct solutions should be considered valid regardless of human expectations, current training paradigms systematically inject human biases about tool selection. This occurs inevitably through cold-start phases where synthetic data generation relies on human-designed demonstrations, creating imitation learning dynamics that persist even after reinforcement learning begins. The history of AI suggests that methods leveraging greater computation for autonomous discovery will ultimately outperform those constrained by human intuitions about appropriate tool usage. However, practical limitations force this bias injection—open-source agents employing cold-start supervised fine-tuning require initial human demonstrations because smaller models lack the capability to explore effectively from scratch. Approaches like process reward models \citep{lightman2023let,wang2025visualprm} likely amplify rather than resolve these human-centric biases. This creates a computational scaling challenge: as systems grow more capable, they should increasingly discover tool-use strategies aligned with their computational strengths rather than human cognitive patterns, yet current training paradigms embed human preferences that may limit this autonomous discovery.

\section{Related Work}
\label{sec:related_work}

\header{Multimodal language models.}
Recently, there have been many advances on open-source multimodal models~\cite{awadalla2023openflamingo, chen2023internvl,  liu2023llava, liu2023improvedllava, liu2024llavanext, dai2024instructblip, li2022blip, li2023blip2, deitke2024molmo}. 
These efforts include training multimodal models to take in multiple images, engage in multi-turn conversations, and even understand videos \cite{liu2024llavanext, jiang2024mantis, li2024llava-ov}. For example, LLaVA-Next achieves strong multi-image understanding through large-scale interleaved visual instruction tuning with M4-Instruct \cite{liu2024llavanext}. 
Similarly, Mantis introduces a new large-scale multi-image instruction tuning dataset Mantis-Instruct for multi-image training \cite{jiang2024mantis}. These efforts pave the foundation for our work on learning vision-language models with image-text interleaved reasoning traces.
over perceptual outputs from vision specialists. 

\header{Multimodal tool-use.}
Recently, there has been increasing interest in enhancing multimodal language models with improved tool-use capabilities~\cite{liu2023llavapluslearningusetools, qi2024cogcomtrainlargevisionlanguage, shao2024visualcot}. LLaVa-Plus demonstrated the feasibility of training such models to utilize visual tools~\cite{liu2023llavapluslearningusetools}. Visual Program Distillation transfers tool-use and reasoning skills into a multimodal model using chain-of-thought (CoT) data derived from programs~\cite{hu2024visual}. Similarly, Visual CoT introduces a synthetic CoT dataset aimed at boosting the reasoning abilities of multimodal models~\cite{shao2024visualcot}. More recently, LLaVa-CoT incorporates both perception and reasoning capabilities inspired by GPT-4o~\cite{xu2025llavacotletvisionlanguage}. CogCoM identifies six key manipulation strategies and trains multimodal models with synthetic chain-of-manipulation (CoM) data~\cite{qi2024cogcomtrainlargevisionlanguage}, while TACO contributes 273K multimodal reasoning traces derived from the outputs of 15 visual tools~\cite{ma2024taco}. Recent works also extent multimodal tool-use with RL, enhancing general reasoning capability \citep{wang2025simple,zhang2025thyme,zhao2025pyvision} and visual search capabiltiy \citep{zheng2025deepeyes,chen2025learning,wang2025vgr}.

\header{Multimodal language model reasoning with RL.}
DeepSeek-R1~\citep{guo2025deepseek} has demonstrated that simple rule-based reinforcement learning can effectively induce strong reasoning behaviors.
This R1-style reasoning paradigm has shown success in multimodal language models~\citep{shen2025vlm,li2025videochat,chen2025visrl}.
VLM-R1~\citep{shen2025vlm} applies reward-driven fine-tuning to improve visual reasoning and generalization beyond supervised baselines.
Vision-R1~\citep{zhan2025vision} introduces human-free alignment through vision-guided rewards, while another Vision-R1~\citep{huang2025vision} leverages CoT-style tasks and reward shaping to boost multi-step reasoning.
UI-R1~\citep{lu2025ui} enhances action prediction in GUI agents via GRPO-based RL, achieving strong performance with compact models.
In the video domain, Video-R1~\citep{feng2025video} and VideoChat-R1~\citep{li2025videochat} improves temporal reasoning through T-GRPO and mixed-modality rewards across diverse video types.

\section{Conclusion}
In this work, we explore tool-usage reinforcement learning to enable models to utilize external vision tools for test-time scaling. We propose \method, a training suite designed to integrate tool-usage with RL, enabling models to optimize tool selection and interleaved text-image-tool reasoning through direct interaction and reward feedback. Our experiments across multiple benchmarks demonstrate that \method successfully raises performance beyond base model, achieving results significantly higher than those obtained through supervised fine-tuning and text-only RL. We hope that \method, along with our fully open-sourced code, dataset, and platform, will serve as a valuable resource for the community.

\section*{Acknowledgment}
We would like to thank Jieyu Zhang, Weikai Huang, and Xirui Li for their insightful feedback and support. 

\appendix
\clearpage
\bibliography{aaai2026}

\clearpage

\section{Dataset Construction Details}

We leverage the SAT dataset and the CoTA dataset from TACO for our training.

The SAT \citep{ray2024sat} dataset is a synthetic VQA dataset designed to enhance the spatial reasoning capabilities of MLLMs.

The CoTA dataset from TACO \citep{ma2024taco} is a synthetic dataset comprising Chain-of-Thought-Action data generated by GPT-4o. However, we leverage it solely as a VQA filter for potential tool utilization.

The CoTA data comprises trajectories of thought, action, and observation. We derive QA pairs from this data using Qwen3-30B-A3B. For open-ended questions within this set, we employ Qwen2.5-VL-32B to synthesize multiple-choice options, thereby transforming the entirety of the CoTA data into a MCQA format. Subsequently, we task Qwen2.5-VL-7B with answering all the questions and retain only those that are answered incorrectly. These erroneous examples are more likely to exhibit both correct and incorrect responses during the sampling of rollouts in GRPO training. The resulting dataset is then randomly split into a cold-start dataset and a RL dataset. For the cold-start dataset, we utilize GPT-4.1 as the agent to synthesize tool-integrated reasoning rollouts, filtering out any rollouts that lead to incorrect answers. 

The SAT data is inherently in a MCQA format. We randomly shuffled the answer options and subsequently filtered out data instances for which Qwen2.5-VL-7B provided an incorrect response.

\section{Experiment Setup Details}
\subsection{Benchmark and Dataset Details}
In this paper, we evaluate five multi-modal benchmarks covering diverse visual reasoning capabilities: CV-Bench \citep{tong2024cambrian1}, Blink \citep{fu2024blink}, Blink-Hard \citep{bigverdi2024perception}, MMVP \citep{tong2024eyes}, and a 1,000 randomly selected subset from SAT \citep{ray2024sat}. Our dataset construction incorporates single-image samples from both SAT \citep{ray2024sat} and Taco \citep{ma2024taco}. The following sections provide detailed descriptions of these benchmarks and datasets:

\begin{itemize}[leftmargin=*, itemsep=0pt]
\item \textbf{CV-Bench} \citep{tong2024cambrian1}: A vision-centric benchmark containing 2,638 manually-inspected examples for evaluating MLLMs. It repurposes standard vision benchmarks (ADE20K \citep{zhou2017scene}, COCO \citep{DBLP:journals/corr/LinMBHPRDZ14}, Omni3D \citep{brazil2023omni3d}) to assess fundamental 2D and 3D understanding capabilities including spatial relationships, object counting, depth ordering, and relative distance estimation. Questions are programmatically constructed and manually verified for clarity and accuracy.
\item \textbf{Blink} \citep{fu2024blink}: A benchmark with 3,807 multiple-choice questions reformatting 14 classic computer vision tasks to test fundamental visual perception abilities. Despite humans achieving 95.70\% accuracy, even top models like GPT-4V and Gemini achieve only 51.26\% and 45.72\% accuracy respectively. Blink highlights significant gaps between human-level visual perception and current MLLM capabilities. Specifically, we experiment with relative depth, count and spatial reasoning tasks in Blink.
\item \textbf{Blink-Hard} \citep{bigverdi2024perception}: A challenging benchmark focused on visual perception tasks requiring deeper 3D understanding and spatial reasoning. It evaluates whether MLMs can produce and reason with intermediate representations like depth maps and bounding boxes. The benchmark serves as a testbed for approaches like Perception Tokens that assist models in tackling complex visual reasoning problems.
\item \textbf{MMVP} \citep{tong2024eyes}: Contains 300 questions based on 150 pairs of ``CLIP-blind'' images that appear similar to CLIP models despite clear visual differences. The benchmark categorizes questions across nine visual patterns and reveals systematic shortcomings in MLLMs' visual perception abilities, with even top models showing performance gaps of over 50\% compared to humans.
\item \textbf{MMMU} \citep{yue2023mmmu}: A massive, multi-discipline benchmark requiring expert-level knowledge to evaluate advanced MLLMs. It consists of 11.5K questions spanning six core disciplines from college-level exams and textbooks. MMMU tests perception and reasoning abilities on complex visual materials like diagrams, charts, and chemical formulas.
\item \textbf{MMBench} \citep{MMBench}: A comprehensive benchmark designed to evaluate MLLMs on a wide range of capabilities, from basic perception to advanced reasoning. It features a carefully curated set of  multiple-choice questions covering 20 skill areas. MMBench employs a circular evaluation strategy on a held-out test set to provide a robust assessment of model performance while preventing overfitting.
\item \textbf{MathVista} \citep{lu2024mathvista}: A benchmark focused on visual mathematical reasoning, designed to test MLLMs on a diverse set of 6,141 problems. Sourced from 28 existing math datasets and 3 newly created ones, its questions cover a wide range of tasks, including geometric problems, statistical chart interpretation, and scientific puzzles. MathVista requires models to integrate visual perception, mathematical logic, and multi-step reasoning.
\item \textbf{SAT} \citep{ray2024sat}: Contains 218K question-answer pairs covering 22K synthetic scenes testing both static and dynamic spatial reasoning. Unlike benchmarks focused on static relationships, SAT \citep{ray2024sat} incorporates perspective-taking and egocentric action recognition. Generated using a photo-realistic physics engine, it can be scaled and extended to include new scenarios.
\item \textbf{Taco} \citep{ma2024taco}: A framework and dataset with over 1 million synthetic chains-of-thought-and-action traces (filtered to 293K high-quality examples). TACO \citep{ma2024taco} enables models to perform step-by-step reasoning while invoking external tools (OCR, depth estimation, calculators) to solve complex visual problems, showing performance improvements of up to 15\% on challenging multimodal tasks.
\end{itemize}

\subsection{Training Details}
\begin{figure}[h]
    \centering
    \includegraphics[width=0.4\textwidth]{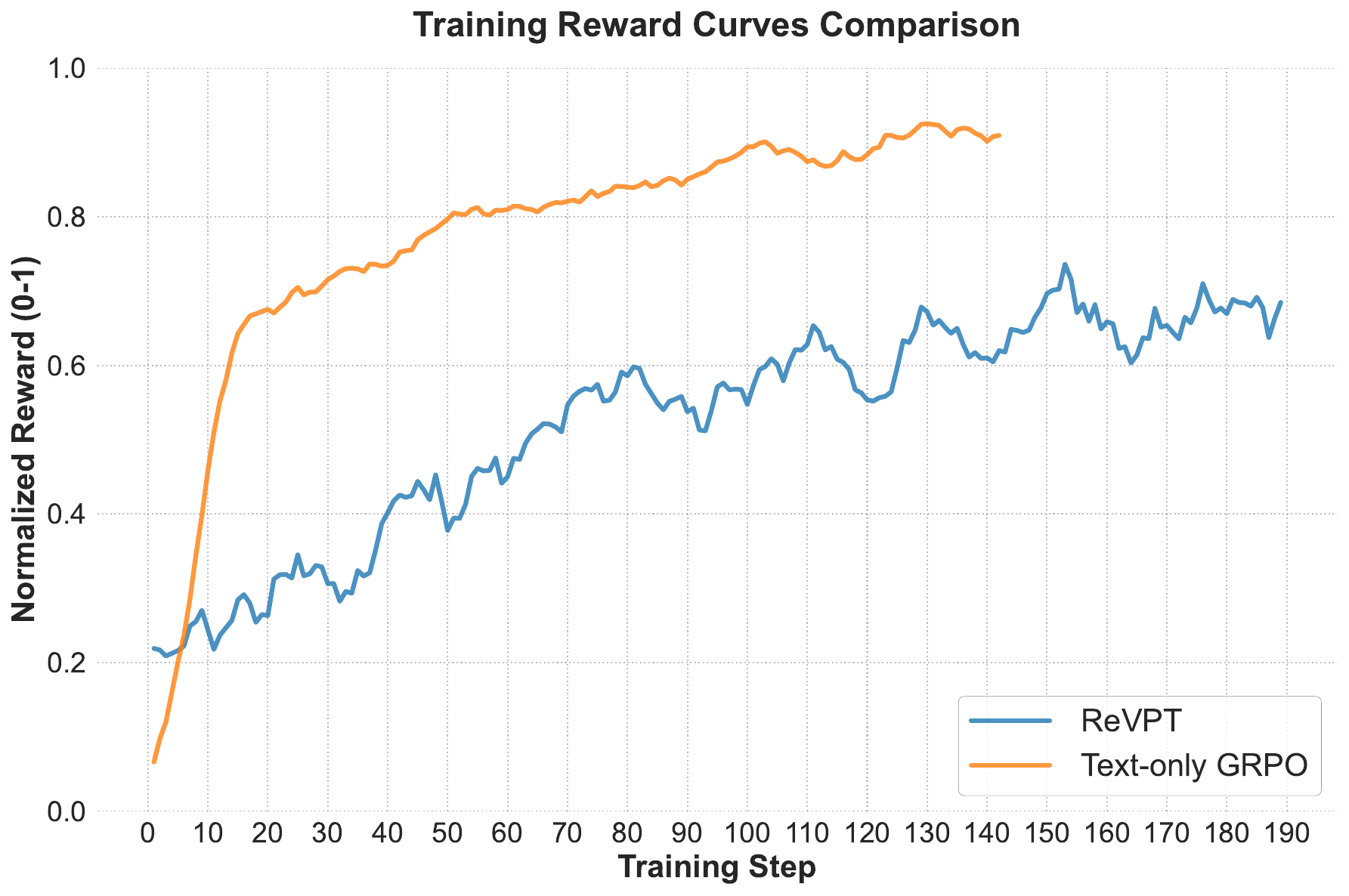}
    \caption{Our reward gradually upgrade and reach converge.}
    \label{fig:loss_curve}
\end{figure}

We fine-tune Qwen2.5-VL-3B and Qwen2.5-VL-7B on eight A800 GPUs. The detailed training parameters for cold-start and reinforcement learning are listed in Table~\ref{tab:sft-hyperparams} and Table~\ref{tab:rl-hyperparams}. The reward curve of our method REVPT compared to text RL training is shown in Figure \ref{fig:loss_curve}.

\begin{table}[H]
\centering
\begin{minipage}[b]{0.45\linewidth}
    \caption{\textbf{Training hyperparameters}}
    \label{tab:sft-hyperparams}
    \resizebox{0.8\columnwidth}{!}{%
        {
        \begin{tabular}{ll} \toprule
        Name                            & Value  \\ \midrule\midrule
        bf16                            & True   \\
        tf32                            & True   \\
        per\_device\_train\_batch\_size & 4      \\
        gradient\_accumulation\_steps   & 2     \\
        lr                              & 1e-5   \\
        weight\_decay                   & 0      \\
        warmup\_ratio                   & 0.1   \\
        lr\_scheduler\_type             & cosine \\
        max\_seq\_len                   & 16384  \\ \bottomrule
        \end{tabular}
        }
    }
\end{minipage}
\begin{minipage}[b]{0.45\linewidth}
    \caption{\textbf{Training hyperparameters}}
    \label{tab:rl-hyperparams}
    \resizebox{0.8\columnwidth}{!}{%
        {
        \begin{tabular}{ll} \toprule
        Name                            & Value  \\ \midrule\midrule
        bf16                            & True   \\
        tf32                            & True   \\
        per\_device\_train\_batch\_size & 8      \\
        ppo\_mini\_batch\_size          & 128     \\
        num\_generation                 & 8      \\
        kl\_loss\_coef                  & 1e-3   \\
        lr                              & 2e-6   \\
        weight\_decay                   & 0.01    \\
        warmup\_ratio                   & 0.03   \\
        lr\_scheduler\_type             & cosine \\
        max\_seq\_len                   & 16384  \\ \bottomrule
        \end{tabular}
        }
    }
\end{minipage}
\end{table}

\begin{table}[H]
\centering

\end{table}

The hyperparameters used for the agent are detailed in Table \ref{tab:agent_hyperparams}. During training with the veRL framework, we observed a degradation in the response quality when batch-inferencing with Qwen2.5-VL by vllm as the batch size increased. Consequently, we imposed a limit on the batch size used for each inference step.

\begin{table}[H]
\centering
\caption{\textbf{Agent config}}
\label{tab:agent_hyperparams}
\resizebox{0.3\columnwidth}{!}{%
{\renewcommand{\arraystretch}{1.0}
\begin{tabular}{ll} \toprule
Name                            & Value  \\ \midrule\midrule
max\_turns                      & 5   \\
max\_tokens\_per\_turn          & 1024   \\
inference\_batch\_size          & 8      \\  \bottomrule
\end{tabular}
}
}
\end{table}

\section{Case Study}
\begin{figure*}[!t]
    \centering
\begin{tcolorbox}[enhanced,attach boxed title to top center={yshift=-3mm,yshifttext=-1mm},boxrule=0.9pt, 
  colback=gray!00,colframe=black!50,colbacktitle=gray,
  boxed title style={size=small,colframe=gray}]

\begin{minipage}{1.00\linewidth}
\textbf{Question}: Considering the relative positions of the bottle (annotated by the red box) and the fork in the image provided, where is the bottle (annotated by the red box) located with respect to the fork? \\
(A) left (B) right\\
\textbf{Answer}: (B)\\
\end{minipage}\hfill

\centering
    \begin{minipage}[b]{0.45\textwidth} 
        \centering
        \includegraphics[width=\linewidth]{figures/edge0.jpg}
    \end{minipage}%
    \begin{minipage}[b]{0.45\textwidth} 
        \centering
        \includegraphics[width=\linewidth]{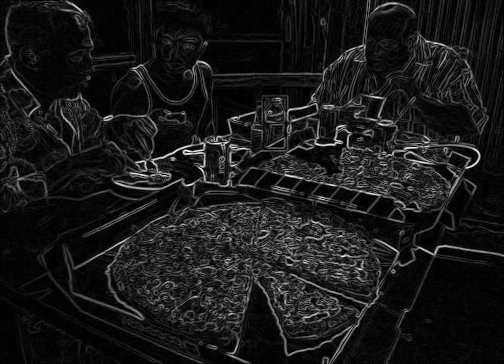}
    \end{minipage}%
\tcblower

\begin{minipage}{1\linewidth}
    \textbf{Assistant:} 
    
    $<$think$>$ bottle (red box) is located on the table near the center of the image. The fork is also visible, likely placed on the table as well. To determine their relative positions, I should analyze the locations of both objects. The bottle appears to be more towards the middle-right side of the table, while the fork is closer to the center or slightly left. To confirm, I will use the edge detection tool to better distinguish the boundaries and positions of these objects.$<$/think$>$
    
    $<$tool\_call$>$\{"name": "edge\_detection", "arguments": \{"image\_id": 0\}\}$<$/tool\_call$>$'\\

\end{minipage}
\begin{minipage}{1\linewidth}
    \textbf{User:} $<$image$>$

    $<$result$>$The edge map for image 0.$<$/result$>$
    \\
\end{minipage}
\begin{minipage}{1\linewidth}
    \textbf{Assistant:} 
    
    $<$think$>$From the edge detection result, it is clear that the bottle (in the red box) is positioned more towards the center-right area of the table. The fork is visible beside the bottle, further to the right. This confirms that the bottle is to the right of the fork.$<$/think$>$
    
    $<$answer$>$Based on the image and the edge detection result, the bottle (red box) is located to the right of the fork. Therefore, the answer is $\backslash$boxed{B}.$<$/answer$>$\\
\end{minipage}

\end{tcolorbox}
\caption{Case study for the tool type: Edge Detection.}
\label{figure: edge}
\end{figure*}

\begin{figure*}[!t]
    \centering
\begin{tcolorbox}[enhanced,attach boxed title to top center={yshift=-3mm,yshifttext=-1mm},boxrule=0.9pt, 
  colback=gray!00,colframe=black!50,colbacktitle=gray,
  boxed title style={size=small,colframe=gray}]

\begin{minipage}{1.00\linewidth}
\textbf{Question}: A bounding box is an annotated rectangle surrounding an object. The edges of bounding boxes should touch the outermost pixels of the object that is being labeled. Given the two bounding boxes on the image, labeled by A and B, which bounding box more accurately localizes and encloses the knife? Select from the following options.\\(A) Box A (B) Box B \\
\textbf{Answer}: (B)\\
\end{minipage}\hfill

\centering
    \begin{minipage}[b]{0.3\textwidth} 
        \centering
        \includegraphics[width=\linewidth]{figures/zoomin0.jpg}
    \end{minipage}%
    \hspace{1pt}
    \begin{minipage}[b]{0.3\textwidth} 
        \centering
        \includegraphics[width=\linewidth]{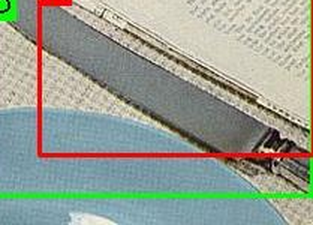}
    \end{minipage}%
\tcblower

\begin{minipage}{1\linewidth}
    \textbf{Assistant:} 
    
    $<$think$>$Let\'s carefully examine the positions of the two bounding boxes, A and B. Box A is around what appears to be a book or a stack of papers, while Box B is around a silver object that resembles a knife. The knife is clearly visible and positioned in Box B, whereas Box A does not contain any knife-like object. To confirm, I\'ll double-check the boundaries of both boxes and their contents.$<$/think$>$
    
    $<$tool\_call$>$\{"name": "zoom\_in", "arguments": \{"image\_id": 0, "bbox": [200, 490, 480, 720], "factor": 1.5\}\}$<$/tool\_call$>$

\end{minipage}
\begin{minipage}{1\linewidth}
    \textbf{User:} $<$image$>$

    $<$result$>$Zoomed image 0 on [200, 490, 480, 720] with 1.5x magnification.$<$/result$>$
    \\
\end{minipage}
\begin{minipage}{1\linewidth}
    \textbf{Assistant:} 
    
    $<$think$>$The zoomed-in view shows the region where both bounding boxes A and B overlap. By closely inspecting this area, I can see that Box B encompasses the handle and blade of a knife, which is not present in Box A. This confirms my initial assessment. Therefore, Box B is the one that correctly encloses the knife.$<$/think$>$
    
    $<$answer$>$The bounding box labeled B more accurately localizes and encloses the knife, as it includes the knife's handle and blade. Therefore, the correct answer is $\backslash$boxed\{B\}.$<$/answer$>$\\
\end{minipage}

\end{tcolorbox}
\caption{Case study for the tool type: Zoom In.}
\label{figure: zoomin}
\end{figure*}

\begin{figure*}[!t]
    \centering
\begin{tcolorbox}[enhanced,attach boxed title to top center={yshift=-3mm,yshifttext=-1mm},boxrule=0.9pt, 
  colback=gray!00,colframe=black!50,colbacktitle=gray,
  boxed title style={size=small,colframe=gray}]

\begin{minipage}{1.00\linewidth}
\textbf{Question}:How many ties are in the image? Select from the following choices.\\
(A) 6 (B) 5 (C) 3 (D) 2 (E) 4 (F) 0 \\
\textbf{Answer}: (E)\\
\end{minipage}\hfill

\centering
    \begin{minipage}[b]{0.45\textwidth} 
        \centering
        \includegraphics[width=\linewidth]{figures/error_object.png}
    \end{minipage}%
    \hspace{1pt}
    \begin{minipage}[b]{0.45\textwidth} 
        \centering
        \includegraphics[width=\linewidth]{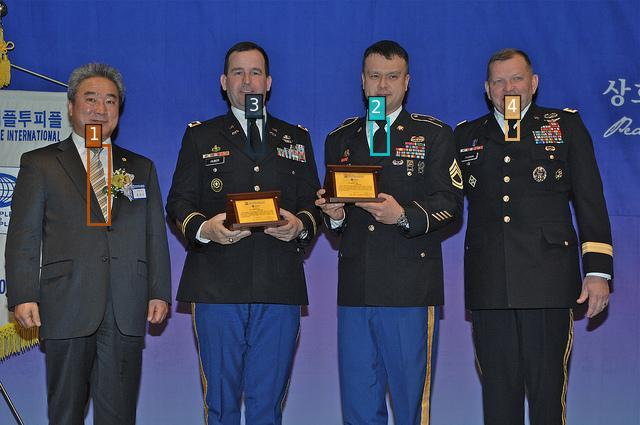}
    \end{minipage}%
\tcblower

\begin{minipage}{1\linewidth}
    \textbf{Assistant:} 
    $<$think$>$To accurately determine the number of ties in the image, it's best to use the object detection tool targeting "tie" to ensure all visible ties are counted, even those partially obscured or at different angles.$<$/think$>$
    
    $<$tool\_call$>$\{"name": "object\_detection", "arguments": \{"image\_id": 0, "objects": ["tie"]\}\}$<$/tool\_call$>$

\end{minipage}
\begin{minipage}{1\linewidth}
    \textbf{User:} $<$image$>$

    $<$result$>$Image 1:Detected 4 object(s) in image 0: 1. tie(0.76): [87, 144, 108, 223] 2. tie(0.46): [370, 117, 387, 153] 3. tie(0.72): [247, 115, 262, 153] 4. tie(0.66): [505, 116, 517, 138]$<$/result$>$
    \\
\end{minipage}
\begin{minipage}{1\linewidth}
    \textbf{Assistant:} 
    $<$think$>$The object detection tool has identified four distinct "ties" in the image, each marked with a bounding box around the neck area. The bounding boxes correspond to the visible neckwear worn by each individual in the image. Since the tool's output matches the visible evidence and there are no other neck accessories present, I can confidently answer the question based on this information.$<$//think$>$

   $<$answer$>$Based on the object detection results and visual confirmation, there are 4 ties in the image. The tool detected one tie for the man on the left (wearing a suit), and three ties for the three men in military uniforms. $\backslash$boxed\{B\}.$<$/answer$>$\\
\end{minipage}

\end{tcolorbox}
\caption{Case study for the tool type: Object Detection.}
\label{figure:Object Detection}
\end{figure*}

\begin{figure*}[!t]
    \centering
\begin{tcolorbox}[enhanced,attach boxed title to top center={yshift=-3mm,yshifttext=-1mm},boxrule=0.9pt, 
  colback=gray!00,colframe=black!50,colbacktitle=gray,
  boxed title style={size=small,colframe=gray}]

\begin{minipage}{1.00\linewidth}
\textbf{Question}:Which object is closer to the camera taking this photo, the table (highlighted by a red box) or the television (highlighted by a blue box)?  Select from the following choices.\\
(A) table (B) television \\
\textbf{Answer}: (E)\\
\end{minipage}\hfill

\centering
    \begin{minipage}[b]{0.45\textwidth} 
        \centering
        \includegraphics[width=\linewidth]{figures/depth0.png}
    \end{minipage}%
    \hspace{1pt}
    \begin{minipage}[b]{0.45\textwidth} 
        \centering
        \includegraphics[width=\linewidth]{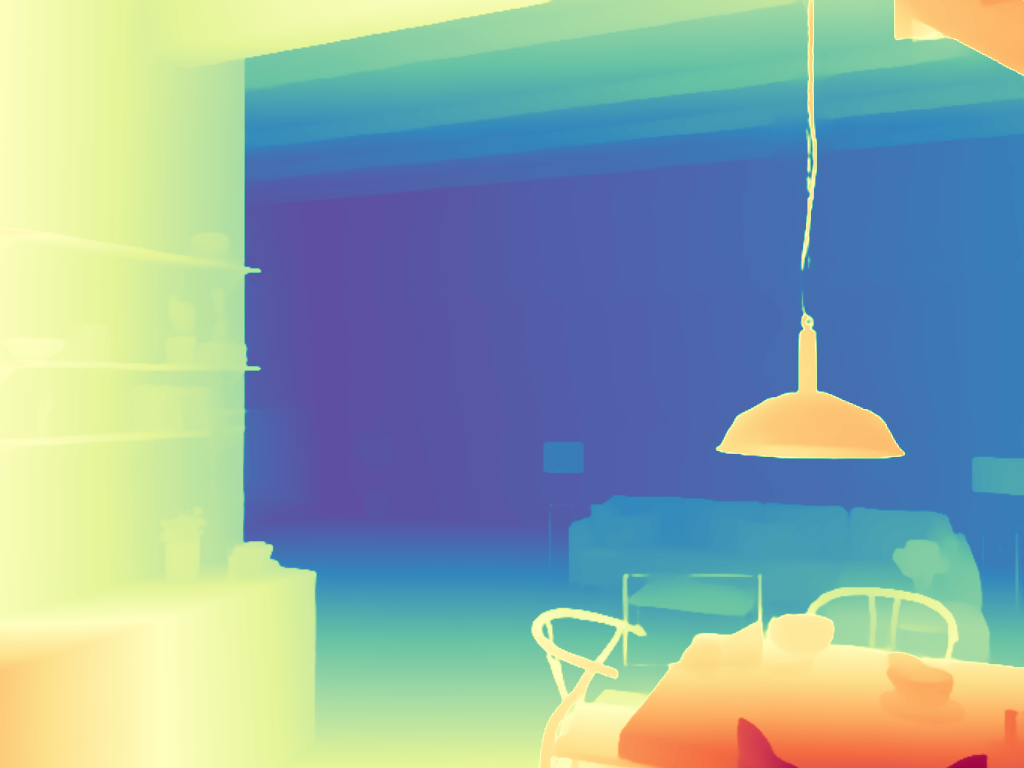}
    \end{minipage}%
\tcblower

\begin{minipage}{1\linewidth}
    \textbf{Assistant:} 
     $<$think$>$To determine which object is closer to the camera.the table (red box) or the television (blue box).it's best to use depth estimation. This will allow me to visually compare the relative distances of these two objects from the camera.$<$//think$>$

    $<$tool\_call$>$\{"name": "depth\_estimation", "arguments": \{"image\_id": 0\}\}$<$/tool\_call$>$

\end{minipage}
\begin{minipage}{1\linewidth}
    \textbf{User:} $<$image$>$

    $<$result$>$Image 1:The colored depth map for image 0.$<$/result$>$
    \\
\end{minipage}
\begin{minipage}{1\linewidth}
    \textbf{Assistant:}

   $<$think$>$In the depth map, warmer colors indicate  areas closer to the camera, while cooler colors represent regions farther away. The area corresponding to the table (red box in the original image) appears much warmer compared to the region where the television (blue box) is located. This suggests that the table is closer to the camera than the television. No further tools are needed as the depth map provides clear evidence.$<$//think$>$

   $<$answer$>$By analyzing the depth map, it is evident that the table (red box) is closer to the camera than the television (blue box). The table's region shows significantly warmer colors, indicating proximity, while the television's area is cooler, signifying it is farther away. //boxed\{A\}. $<$//answer$>$\\
\end{minipage}

\end{tcolorbox}
\caption{Case study for the tool type: Depth Estimation.}
\label{figure: depth}
\end{figure*}

\begin{figure*}[!t]
    \centering
\begin{tcolorbox}[enhanced,attach boxed title to top center={yshift=-3mm,yshifttext=-1mm},boxrule=0.9pt, 
  colback=gray!00,colframe=black!50,colbacktitle=gray,
  boxed title style={size=small,colframe=gray}]

\begin{minipage}{1.00\linewidth}
\textbf{Question}: Two points are circled on the image, labeled by A and B beside each circle. Which point is closer to the camera? Select from the following choices.\\
(A) A is closer (B) B is closer \\
\textbf{Answer}: (B)\\
\end{minipage}\hfill

\centering
    \begin{minipage}[b]{0.45\textwidth} 
        \centering
        \includegraphics[width=\linewidth]{figures/error_depth.jpg}
    \end{minipage}%
    \begin{minipage}[b]{0.45\textwidth} 
        \centering
        \includegraphics[width=\linewidth]{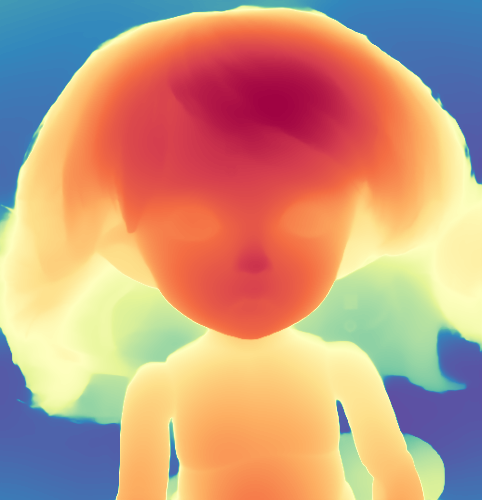}
    \end{minipage}%
\tcblower

\begin{minipage}{1\linewidth}
    \textbf{Assistant:} 
    
    $<$think$>$Looking at the image, both points A and B are marked near the face of the doll. To determine which one is closer to the camera, I need to analyze their positions relative to the viewer\'s perspective. Point A appears lower on the doll\'s face compared to point B, which is higher up on the forehead. However, this could be due to the camera angle or perspective. To be more certain, I should use depth estimation to compare the actual spatial distances of points A and B from the camera.$<$/think$>$
    
    $<$tool\_call$>$\{"name": "depth\_estimation", "arguments": \{"image\_id": 0\}\}$<$/tool\_call$>$

\end{minipage}
\begin{minipage}{1\linewidth}
    \textbf{User:} $<$image$>$

    $<$result$>$Image 1:The colored depth map for image 0.$<$/result$>$
    \\
\end{minipage}
\begin{minipage}{1\linewidth}
    \textbf{Assistant:} 
    
    $<$think$>$In the depth map, warmer colors indicate areas that are closer to the camera, while cooler colors represent regions farther away. Observing the map, the region around point A (near the lower part of the face) appears much warmer compared to the area around point B (higher on the forehead). This confirms that point A is indeed closer to the camera than point B.$<$/think$>$
    
    $<$answer$>$Based on the depth map, point A is closer to the camera than point B. The warm color at point A supports this conclusion. $\backslash$boxed\{A\}$<$/answer$>$\\
\end{minipage}

\end{tcolorbox}
\caption{Case study for the error: Model's Misunderstanding  of Tool Results (Depth Estimation).}
\label{figure: error_depth}
\end{figure*}

We show some case studies that our model successfully using tools in Figure \ref{figure: edge} ,\ref{figure: zoomin} ,\ref{figure:Object Detection} and \ref{figure: depth}. We also include some failure cases that visual tools return wrong output and disturb model's reasoning chain in  \ref{figure: error_obj}, \ref{figure: error_depth} and \ref{figure: error_obj2}.

\begin{figure*}[!t]
    \centering
\begin{tcolorbox}[enhanced,attach boxed title to top center={yshift=-3mm,yshifttext=-1mm},boxrule=0.9pt, 
  colback=gray!00,colframe=black!50,colbacktitle=gray,
  boxed title style={size=small,colframe=gray}]

\begin{minipage}{1.00\linewidth}
\textbf{Question}:How many pictures are in the image??  Select from the following choices.\\
(A) 0  (B) 5  (C) 2  (D) 1  (E) 3   (F) 4 \\
\textbf{Answer}: (E)\\
\end{minipage}\hfill

\centering
    \begin{minipage}[b]{0.45\textwidth} 
        \centering
        \includegraphics[width=\linewidth]{figures/toolr.png}
    \end{minipage}%
    \hspace{1pt}
    \begin{minipage}[b]{0.45\textwidth} 
        \centering
        \includegraphics[width=\linewidth]{figures/toolr1.png}
    \end{minipage}%
\tcblower

\begin{minipage}{1\linewidth}
    \textbf{Assistant:} 
    $<$think$>$ To accurately determine the number of pictures in the image, I should use the object detection tool to identify all the pictures present. This will help ensure I don't miss any that may not be immediately obvious.$<$//think$>$
    
    $<$tool\_call$>$\{"name": "object\_detection", "arguments": \{"image\_id": 0, "objects": ["picture"]\}\} $<$/tool\_call$>$

\end{minipage}
\begin{minipage}{1\linewidth}
    \textbf{User:} $<$image$>$

    $<$result$>$Image 1:Detected 3 object(s) in image 0: 1. picture(0.76): [483, 144, 570, 279] 2. picture(0.72): [115, 158, 188, 247] 3. picture(0.42): [747, 194, 781, 280]  $<$/result$>$
    \\
\end{minipage}
\begin{minipage}{1\linewidth}
    \textbf{Assistant:}

   $<$think$>$The object detection tool has identified three pictures in the image, each marked with a bounding box and label. The bounding boxes correspond to the framed objects on the wall: one on the left above the bookshelf, one in the middle above the cabinet, and one on the right wall. Since the detection confidence is high and matches the visible frames, there is no need for further tool usage. I am ready to answer the question based on this information.$<$/think$>$

   $<$answer$>$Based on the object detection results, there are three pictures in the image. The bounding boxes and labels confirm this count. $\backslash$boxed\{C\}. $<$/answer$>$\\
\end{minipage}

\end{tcolorbox}
\caption{Case study for the error:Model's Misunderstanding  of Tool Results (Object Detection).}
\label{figure: error_obj}
\end{figure*}

\begin{figure*}[!t]
    \centering
\begin{tcolorbox}[enhanced,attach boxed title to top center={yshift=-3mm,yshifttext=-1mm},boxrule=0.9pt, 
  colback=gray!00,colframe=black!50,colbacktitle=gray,
  boxed title style={size=small,colframe=gray}]

\begin{minipage}{1.00\linewidth}
\textbf{Question}: How many pillows are in the image? Select from the following choices.\\
(A) 2  (B) 0  (C) 1  (D) 3 \\
\textbf{Answer}: (B)\\
\end{minipage}\hfill

\centering
    \begin{minipage}[b]{0.45\textwidth} 
        \centering
        \includegraphics[width=\linewidth]{figures/56_0.png}
    
    \end{minipage}%
\tcblower

\begin{minipage}{1\linewidth}
    \textbf{Assistant:} 
    
    $<$think$>$To accurately determine the number of pillows in the image, it's best to use object detection targeting \"pillow\" or similar objects. This will help ensure I don't miss any pillows that may not be immediately obvious.$<$/think$>$
    
    $<$tool\_call$>$\{"name": "object\_detection", "arguments": \{"image\_id": 0, "objects": ["pillow"]\}\}$<$/tool\_call$>$

\end{minipage}
\begin{minipage}{1\linewidth}
    \textbf{User:} 
    $<$result$>$No objects matching 'pillow.' detected in image 0.$<$/result$>$
    \\
\end{minipage}
\begin{minipage}{1\linewidth}
    \textbf{Assistant:} 
    
    $<$think$>$The object detection tool did not find any pillows, but it is not always reliable. Visually inspecting the image, there appears to be one pillow visible on the bed in the background. The rest of the items look like bags, clothes, or bedding. Since the detection tool missed it, and only one pillow is clearly visible, it seems reasonable to conclude that there is only one pillow in the image.$<$/think$>$
    
    $<$answer$>$After careful inspection and considering the output of the object detection tool, there is only one visible pillow in the image.$\backslash$boxed\{C\}$<$/answer$>$\\
\end{minipage}

\end{tcolorbox}
\caption{Case study for the error: Flawed Tool Results (Object Detection).}
\label{figure: error_obj2}
\end{figure*}

\end{document}